\documentclass[sigconf]{acmart}

\usepackage{graphicx}
\usepackage{algorithm}
\usepackage{algorithmic}
\usepackage{subfigure}
\usepackage{array}
\usepackage{setspace}
\usepackage{float}
\usepackage{url}
\usepackage{balance}
\usepackage{amsfonts}
\usepackage{amsmath}
\usepackage{rotating}
\usepackage{multirow}
\usepackage{color}
\usepackage{bm}
\usepackage{enumitem}
\copyrightyear{2020}
\acmYear{2020}
\setcopyright{acmcopyright}\acmConference[SIGIR '20]{Proceedings of the 43rd International ACM SIGIR Conference on Research and Development in Information Retrieval}{July 25--30, 2020}{Virtual Event, China}
\acmBooktitle{Proceedings of the 43rd International ACM SIGIR Conference on Research and Development in Information Retrieval (SIGIR '20), July 25--30, 2020, Virtual Event, China}
\acmPrice{15.00}
\acmDOI{10.1145/3397271.3401168}
\acmISBN{978-1-4503-8016-4/20/07}

\begin{document}
\title{Minimally Supervised Categorization of Text with Metadata}

\author{Yu Zhang$^{1*}$, Yu Meng$^{1*}$, Jiaxin Huang$^{1}$, Frank F. Xu$^{2}$, Xuan Wang$^{1}$, Jiawei Han$^{1}$}
\affiliation{
\institution{$^1$Department of Computer Science, University of Illinois at Urbana-Champaign, IL, USA} 
\institution{$^2$Language Technologies Institute, Carnegie Mellon University, PA, USA}
\institution{$^{1}$\{yuz9, yumeng5, jiaxinh3, xwang174, hanj\}@illinois.edu, \ \ \ $^2$frankxu@cmu.edu}
}
\thanks{$^*$Equal Contribution.}

\begin{abstract}
Document categorization, which aims to assign a topic label to each document, plays a fundamental role in a wide variety of applications. Despite the success of existing studies in conventional supervised document classification, they are less concerned with two real problems: (1) \textit{the presence of metadata}: in many domains, text is accompanied by various additional information such as authors and tags. Such metadata serve as compelling topic indicators and should be leveraged into the categorization framework; (2) \textit{label scarcity}: labeled training samples are expensive to obtain in some cases, where categorization needs to be performed using only a small set of annotated data. In recognition of these two challenges, we propose \textsc{MetaCat}, a minimally supervised framework to categorize text with metadata. Specifically, we develop a generative process describing the relationships between words, documents, labels, and metadata. Guided by the generative model, we embed text and metadata into the same semantic space to encode heterogeneous signals. Then, based on the same generative process, we synthesize training samples to address the bottleneck of label scarcity. We conduct a thorough evaluation on a wide range of datasets. Experimental results prove the effectiveness of \textsc{MetaCat} over many competitive baselines.
\end{abstract}




\maketitle

\begin{spacing}{1}
\section{Introduction}
Our daily life is surrounded by a wealth of text data, ranging from news articles to social media and scientific publications. Document categorization
(i.e., assigning a topic label to each document) serves as a critical first step towards organizing, searching and analyzing such a vast spectrum of text data. Many real applications, such as sentiment analysis \cite{wang2018sentiment} and location prediction \cite{cheng2010you}, can also be cast as a document categorization task.

Although deep neural models \cite{kim2014convolutional,yang2016hierarchical} equipped with word embeddings \cite{mikolov2013distributed} and pre-trained language models \cite{devlin2019bert} have achieved superior performance on document categorization, existing studies are less concerned with two problems in real applications: (1) \textit{The presence of metadata}: metadata prevalently exists in many data sources, especially social media platforms. For example, each tweet is associated with a Twitter user (i.e., its creator) and several hashtags; each Amazon/Yelp review has its product and user information. 
Such metadata makes each ``document'' a complex object beyond plain text. As a result, heterogeneous signals should be leveraged in the categorization process. (2) \textit{Label scarcity}: conventional supervised text classification methods \cite{kim2014convolutional,yang2016hierarchical,liu2017deep,sun2019fine} rely on a sufficient number of labeled documents as training data. However, annotating enough data for training classification models could be expensive. Moreover, manual labeling requires technical expertise in some domains (e.g., arXiv papers and GitHub repositories), which incurs additional cost. In these cases, it would be favorable to perform categorization using only a small set of training samples that an individual user can afford to provide.

\begin{figure*}
\centering
\subfigure[\textsc{GitHub Repository}]{
\includegraphics[height=4.05cm]{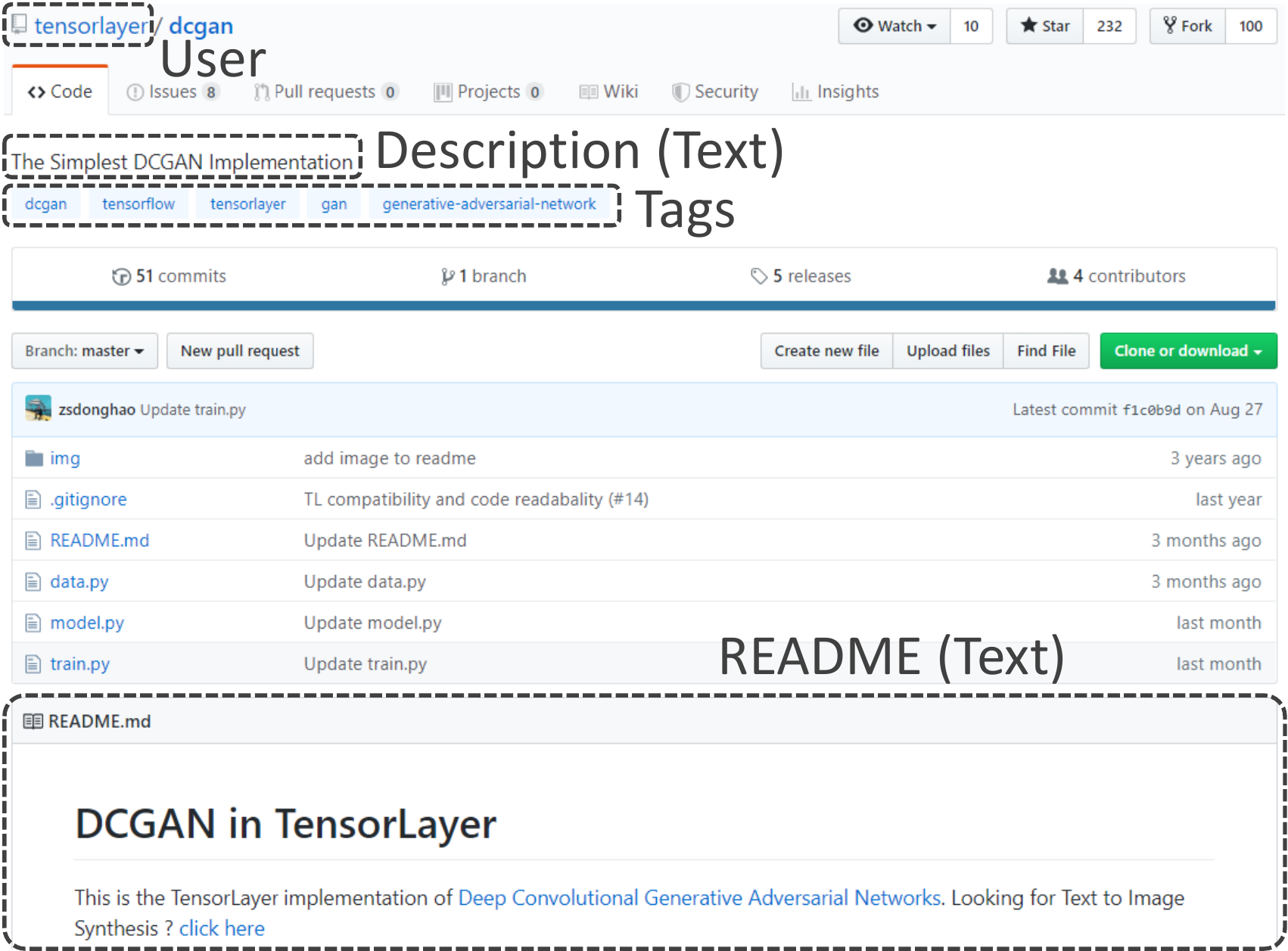}}
\hspace{-0.5ex}
\subfigure[\textsc{Tweet}]{
\includegraphics[height=4.05cm]{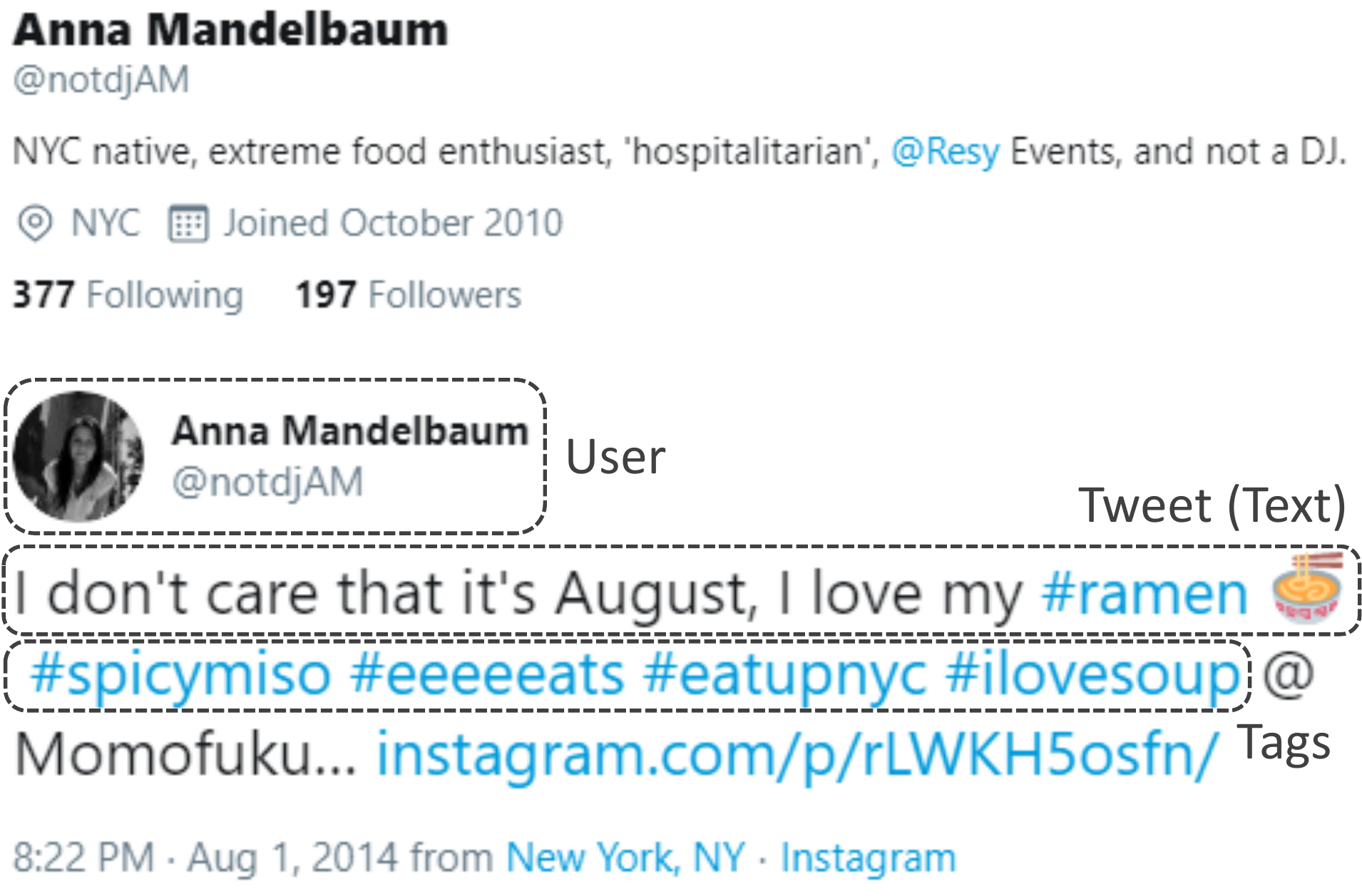}}
\hspace{-0.5ex}
\subfigure[\textsc{Amazon Review}]{
\includegraphics[height=4.05cm]{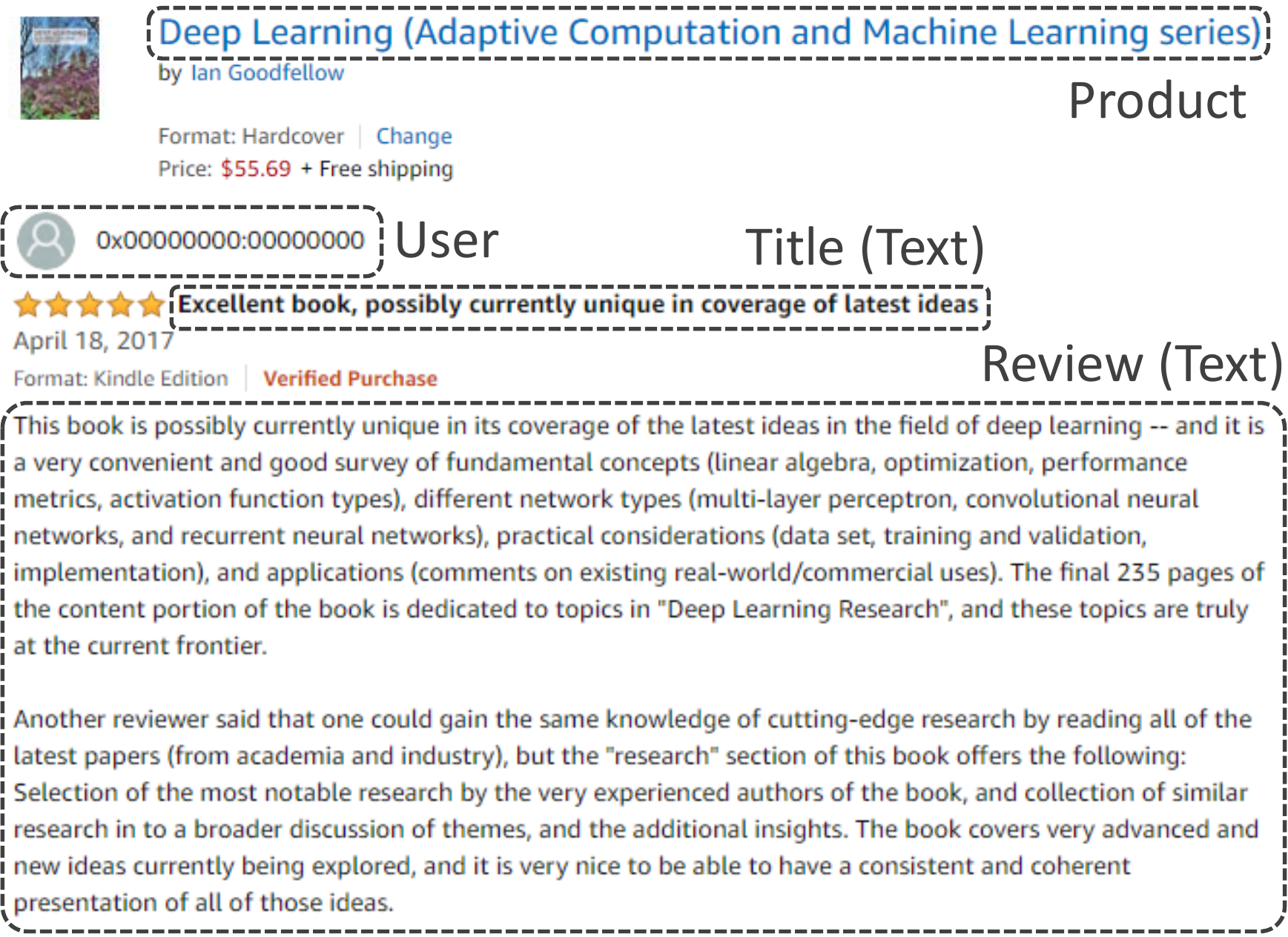}}
\vspace{-0.5em}
\caption{Three examples of documents with metadata.} 
\vspace{-0.5em}
\label{fig:example}
\end{figure*}

Combining the statements above, we define our task as \textit{minimally supervised categorization of text with metadata}. There exist previous attempts that incorporate \textit{metadata} in text categorization \cite{tang2015learning,rosen2004author,kim2019categorical}. For example, Kim et al. \cite{kim2019categorical} customize embeddings, transformation matrices and encoder weights in the neural classifier according to metadata information. While it does improve classification accuracy, the model is designed under the fully supervised setting and requires massive training data. Along another line of work, researchers focus on text classification under \textit{weak supervision}\footnote{\textit{Weak supervision} implies that less than a dozen labeled documents are provided for each category.} \cite{meng2018weakly,xiao2019efficient}. However, they still view documents as plain text sequences and thus are not optimized under the presence of metadata. 


In order to effectively leverage multi-modal signals and scarce labeled data \textit{jointly}, in this paper, we propose a unified, embedding-based categorization framework called \textsc{MetaCat}. The design of \textsc{MetaCat} contains two key ideas: (1) to deal with data heterogeneity, we embed words, documents, labels and various metadata (e.g., users, tags and products) into the same latent space to characterize their relationships; (2) to tackle label scarcity, we generate synthesized training samples based on the learned text and metadata embeddings. The generated data, together with the ``real'' training data, are used to train a neural classifier. We propose a generative model to simultaneously facilitate these two key ideas.
Based on the generative process, our model (1) learns the embedding vectors of all elements via maximum likelihood estimation on the textual and metadata statistics;
(2) synthesizes training samples with both text and metadata.

We conduct experiments on five real-world datasets from three different domains (GitHub repositories, tweets and Amazon reviews). The results reveal that \textsc{MetaCat} outperforms various text-based and graph-based benchmark approaches. Moreover, we show (1) the superiority of our embedding module towards heterogeneous network embedding methods \cite{shang2016meta,dong2017metapath2vec,fu2017hin2vec} and (2) the significant contribution of our generation module under weak supervision.

To summarize, this paper makes the following contributions:
\begin{itemize}[leftmargin=*]
    \item We formulate the problem of minimally supervised categorization of text with metadata. It poses two unique challenges: data heterogeneity and label scarcity. 
    \item We propose a principled generative process to characterize relationships between words, documents, labels and various metadata. 
    \item We develop a novel framework \textsc{MetaCat} with an embedding module and a generation module to tackle the two challenges, respectively. Both modules are derived from the proposed generative process. We also provide theoretical interpretations of our design based on a spherical probabilistic model.
    \item We conduct extensive experiments to demonstrate the effectiveness of \textsc{MetaCat} and verify the contribution of both embedding and generation modules.
\end{itemize}

\section{Preliminaries}
\subsection{Problem Definition}
Given a collection of documents $\mathcal{D}=\{d_1,...,d_{|\mathcal{D}|}\}$ and a label space $\mathcal{L}=\{l_1,...,l_{|\mathcal{L}|}\}$, text classification aims to assign a class label $l_i$ to each document $d_j$. To characterize each class $l \in \mathcal{L}$, a set of labeled documents $\mathcal{D}_l \subseteq \mathcal{D}$ is provided as training data. Our problem is different from many previous text classification studies \cite{kim2014convolutional,tang2015pte,yang2016hierarchical} from two perspectives. First, each document $d_i$ is accompanied by some metadata $m_i$ (which will be discussed in Section \ref{sec:tme}). Second, traditional supervised methods require sufficient annotated documents (i.e., $|\mathcal{D}_l|$ is large), while we rely on minimal supervision where $|\mathcal{D}_l|$ is small (e.g., less than a dozen). Formally, we define the problem as follows.

\begin{definition}
	\textsc{(Problem Definition)} Given documents $\mathcal{D}$, metadata $\mathcal{M}$, label space $\mathcal{L}$ and a small set of training data $\{\mathcal{D}_l:l\in \mathcal{L}\}$, the task is to assign a label $l_i \in \mathcal{L}$ to each document $d_j \in \mathcal{D}$.
\end{definition}

\subsection{Text and Metadata}
\label{sec:tme}
The presence of metadata is common in social media corpora. Let us start from some concrete examples.
Figure \ref{fig:example} shows three ``documents'' extracted from different social media platforms. They contain various types of information, which can be grouped as follows.

\vspace{1mm}

\noindent \textbf{Text.} Textual information is the main body of a document. Some data may have multiple segments of text (e.g., Description+README in GitHub repositories, and Title+Review in Amazon reviews). To simplify our discussion, we concatenate them together for further utilization.


\vspace{1mm}

\noindent \textbf{User/Author.} The author of a document is a strong topic indicator because one person often has consistent interests. For example, in Figure \ref{fig:example}(a), the user ``\textit{tensorlayer}'' publishes repositories mostly on deep learning; in Figure \ref{fig:example}(b), we may infer that the user have many food-related tweets based on her self introduction.

\vspace{1mm}

\noindent \textbf{Tag/Hashtag.} Tags are a set of concepts describing the documents. Although more concise than text, they still suffer from noise and sparsity. For example, many hashtags in tweets appear only once, and over 70\% of GitHub repositories in our collections have no tags. The major difference between words and tags is that words have orders in a sentence (thus local context information needs to be modeled), while tags are swappable. 

\vspace{1mm}

\noindent \textbf{Product.} Product name information is distinctive in Amazon reviews. It is related to both the topic and the sentiment of reviews.

As we can see, the types of metadata are diverse. One can give even more varieties.
However, in most cases, topic-indicative metadata can be divided into two major categories according to their relationships with documents.

\vspace{1mm}

\noindent \textbf{Global Metadata.} Global metadata ``causes'' the generation of documents. The semantics of a document is based on the semantics of its global metadata. For example, there first exists a user or a product, then some reviews are created by the user/for the product. This generative order cannot be reversed. Therefore, users and products are both global.

\vspace{1mm}

\noindent \textbf{Local Metadata.} Local metadata ``describes'' the overall idea of a document. The semantics of local metadata is based on the semantics of its associated document. From this perspective, tags/hashtags are local. We can also say words in text are local, although they are beyond our discussion of ``metadata''. 

\subsection{The Von Mises-Fisher Distribution}
\newcommand{\bmmu}{{\bm \mu}}
\newcommand{\bmx}{{\bm x}}
The von Mises-Fisher (vMF) distribution defines a probability density over points on a unit sphere. It is parameterized by a mean direction vector $\bmmu$ and a concentration parameter $\kappa$. Let $\mathbb{S}^{p-1} = \{\bmx \in \mathbb{R}^p:||\bmx||_2=1\}$ denote the $p$-dimensional unit sphere. Then the probability density function of $\text{vMF}_p(\bmmu,\kappa)$ is defined as
$$
f_{\rm vMF}(\bmx;\bmmu,\kappa) = c_p(\kappa)\exp(\kappa \bmx^T\bmmu), \ \ \bmx \in \mathbb{S}^{p-1},
$$
where $||\bmmu||_2=1$ and $\kappa > 0$. The normalization constant $c_p(\kappa)$ is given by 
$$
c_p(\kappa) = \frac{\kappa^{p/2-1}}{(2\pi)^{p/2}I_{p/2-1}(\kappa)},
$$
where $I_r(\cdot)$ represents the modified Bessel function of the first kind at order $r$ \cite{mardia2009directional,gopal2014mises}.
Intuitively, the vMF distribution can be interpreted as a normal distribution on the sphere. It characterizes data points concentrating around the mean direction $\bmmu$, and they are more concentrated if $\kappa$ is large.

\section{Method}
\begin{figure}
\centering
\subfigure[\textsc{GitHub / Tweet}]{
\includegraphics[height=3.3cm]{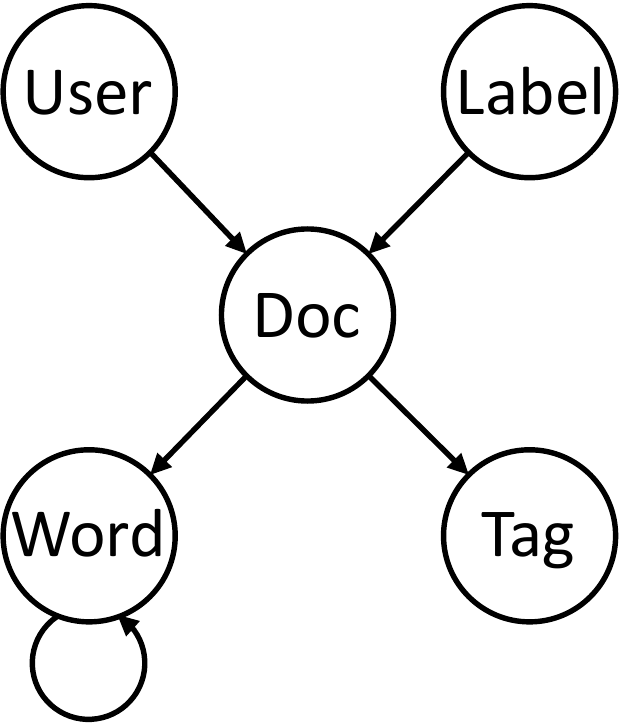}}
\hspace{1ex}
\subfigure[\textsc{The General Case}]{
\includegraphics[height=3.3cm]{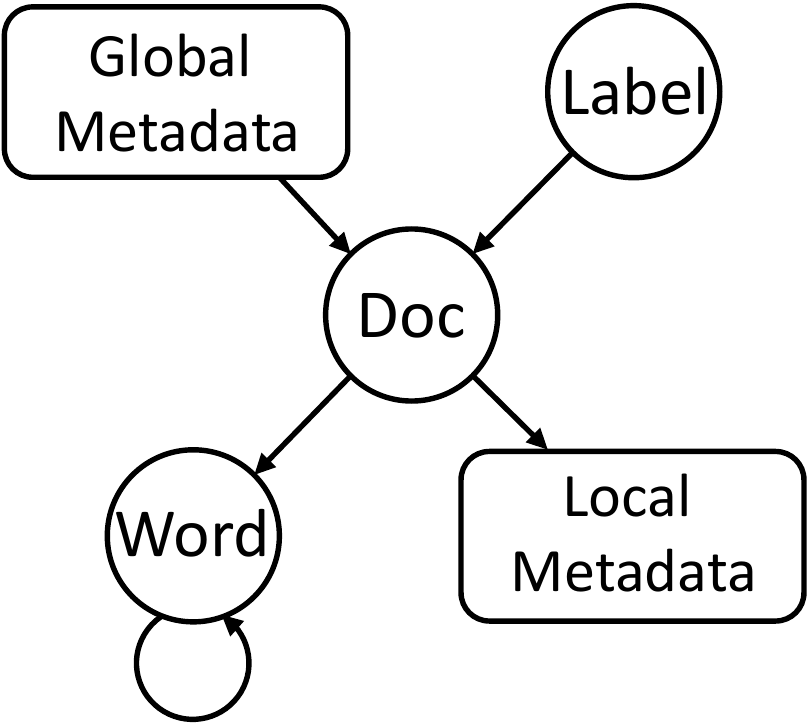}}
\vspace{-1em}
\caption{The generative process of text and metadata. The self loop of ``Word'' represents the step of words generating contexts.} 
\vspace{-1.5em}
\label{fig:story}
\end{figure}

\textsc{MetaCat} consists of two key modules: embedding learning and training data generation, which are proposed to tackle data heterogeneity and label scarcity, respectively. We start this section by introducing a generative process (Figure \ref{fig:story}) that guides the design of these two modules. Following this process, the embedding module learns representation vectors through maximizing the likelihood of observing all data, and the generation module synthesizes training documents with both text and metadata.

For the sake of clarity, we use a specific case (Figure \ref{fig:story}(a)) to illustrate our framework. This is also the case of GitHub repositories and tweets. In Section \ref{sec:general}, we discuss how to generalize the framework (Figure \ref{fig:story}(b)).

\subsection{The Generative Process}
\label{sec:story}
\newcommand{\bme}{{\bm e}}
According to Figure \ref{fig:story}(a), the generative process can be decomposed into four steps.

\vspace{1mm}

\noindent \textbf{User \& Label $\rightarrow$ Document.} When a user decides to write a document given a topic, s/he first has an overall idea of what to talk about. This ``overall idea'' can be represented as document embedding $\bme_d$, which should be close to user embedding $\bme_u$ and label embedding $\bme_l$. Inspired by the softmax function in word2vec \cite{mikolov2013distributed}, we define the generation probability as follows:
\begin{equation}
p(d|u,l) \propto \exp(\bme_d^T\bme_u) \cdot \exp(\bme_d^T\bme_l).
\label{eqn:ruf}
\end{equation}

\noindent \textbf{Document $\rightarrow$ Word.} Given the overall idea $\bme_d$, we can write down words that are coherent with the meaning of the entire document. To encourage such coherence, we have 
\begin{equation}
p(w|d) \propto \exp(\bme_w^T\bme_d).
\label{eqn:wr}
\end{equation}

\noindent \textbf{Document $\rightarrow$ Tag.} Tags can be generated similarly.
\begin{equation}
p(t|d) \propto \exp(\bme_t^T\bme_d).
\label{eqn:tr}
\end{equation}

\noindent \textbf{Word $\rightarrow$ Context.} Different from tags, words in text carry sequential information. Tang et al. \cite{tang2015pte} point out that the embedding of a word is related to not only its global document representation (i.e., $\bme_d$) but also its local context. To be specific, given a sequence of words $w_1w_2...w_n$, the local context of $w_i$ is defined as $\mathcal{C}(w_i, h) = \{w_j : i-h \leq j\leq i+h, i\neq j\}$, where $h$ is the context window size. Mikolov et al. \cite{mikolov2013distributed} propose the Skip-Gram model in which $\mathcal{C}(w_i, h)$ is predicted given the center word $w_i$. Following \cite{mikolov2013distributed}, we define the generation probability to be
\begin{equation}
p(\mathcal{C}(w_i, h)|w_i) \propto \prod_{w_j \in \mathcal{C}(w_i, h)}\exp(\bme'^{\ T}_{w_j} \bme_{w_i}).
\label{eqn:wcf}
\end{equation}
Note that each word $w$ has two embeddings: $\bme_w$ when $w$ is viewed as a center word and $\bme'_w$ when $w$ is a context word \cite{mikolov2013distributed}.

\vspace{1mm}

\noindent \textbf{Connections to the vMF Distribution.} Now we explain how these conditional probabilities are related to the vMF distribution. Taking $p(w|d)$ as an example, according to Eq. (\ref{eqn:wr}), we know that
\begin{equation}
  p(w|d) = \frac{\exp(\bme_w^T\bme_d)}{\sum_{w' \in \mathcal{V}}\exp(\bme_{w'}^T\bme_d)}, 
  \label{eqn:disc}
\end{equation}
where $\mathcal{V}$ is the vocabulary.
Following \cite{meng2020cate}, when $|\mathcal{V}|$ goes to infinity and the representation vectors of all elements are assumed to be \textit{unit vectors}, we can generalize Eq. (\ref{eqn:disc}) to the continuous case:
\begin{equation}
\lim_{|\mathcal{V}|\rightarrow \infty} p(w|d) = \frac{\exp(\bme_w^T\bme_d)}{\int_{\mathbb{S}^{p-1}}\exp(\bme_{w'}^T\bme_d) d\bme_{w'}}.
\label{eqn:lim1}
\end{equation}
To calculate the denominator, one needs to note that the probability density function of vMF distribution integrates to 1 over the whole sphere. Therefore,
\begin{equation}
1=\int_{\mathbb{S}^{p-1}}f_{\rm vMF}(\bme_{w'}; \bme_d, 1)d\bme_{w'} = c_p(1)\int_{\mathbb{S}^{p-1}}\exp(\bme_{w'}^T\bme_d) d\bme_{w'}.
\label{eqn:int}
\end{equation}
Combining Eqs. (\ref{eqn:lim1}) and (\ref{eqn:int}), we get
\begin{equation}
\lim_{|\mathcal{V}|\rightarrow \infty} p(w|d) = c_p(1)\exp(\bme_w^T\bme_d) = \text{vMF}_p(\bme_d, 1).
\label{eqn:vmf0}
\end{equation}
Similarly,
\begin{equation}
\begin{split}
&\lim_{|\mathcal{D}|\rightarrow \infty} p(d|u,l) \propto \text{vMF}_p(\bme_u, 1)\cdot \text{vMF}_p(\bme_l, 1), \\
&\lim_{|\mathcal{T}|\rightarrow \infty} p(t|d) = \text{vMF}_p(\bme_d, 1), \\
&\lim_{|\mathcal{V}|\rightarrow \infty} p(w_j|w_i) = \text{vMF}_p(\bme_{w_i}, 1),
\end{split}
\label{eqn:vmf}
\end{equation}
where $\mathcal{T}$ is the set of tags appearing in the corpus.

The probability $p(d|u,l)$ needs to be elaborated more here. During our embedding step, $l$ is unknown in many cases because only a small proportion of documents have label information. When $l$ is missing, it is natural to assume that $\bme_l$ can be any vector on the sphere with equal probability (i.e., $\bme_l \sim U(\mathbb{S}^{p-1})$). In this case, Eq. (\ref{eqn:ruf}) becomes
\begin{equation}
\begin{split}
p(d|u,l) &\propto \mathbb{E}_{\bme_l \sim U(\mathbb{S}^{p-1})}[\exp(\bme_d^T\bme_u) \cdot \exp(\bme_d^T\bme_l)] \\
&= \exp(\bme_d^T\bme_u) \cdot \mathbb{E}_{\bme_l \sim U(\mathbb{S}^{p-1})}[\exp(\bme_d^T\bme_l)].
\label{eqn:nol}
\end{split}
\end{equation}
For any fixed $\bme_d$, using Eq. (\ref{eqn:int}), we have
\begin{equation}
    \mathbb{E}_{\bme_l \sim U(\mathbb{S}^{p-1})}[\exp(\bme_d^T\bme_l)]
    \propto \int_{\mathbb{S}^{p-1}} \exp(\bme_d^T\bme_l) d\bme_l = 1/c_p(1).
\end{equation}
In other words, this term is the same for any $\bme_d \in \mathbb{S}^{p-1}$. Therefore, when there is no label information,
\begin{equation}
    p(d|u,l) \propto \exp(\bme_d^T\bme_u).
\label{eqn:lx}
\end{equation}

Our assumption that all embeddings are unit vectors has empirical bases because normalizing embeddings onto a sphere is common practice in natural language processing \cite{levy2015improving,xing2015normalized}. 


\subsection{Generation-Guided Embedding Learning} 
\label{sec:embedding}
Given the generative process, we are able to learn $\bme_u$, $\bme_l$, $\bme_d$, $\bme_t$, $\bme_{w}$ and $\bme'_{w}$ through maximum likelihood estimation.

\vspace{1mm}

\noindent \textbf{Likelihood.}
Assume all embedding vectors are parameters of our generative model. 
The likelihood of observing the whole corpus (including metadata) is
\begin{equation}
\begin{split}
\mathcal{J} = & \prod_{d\in \mathcal{D}}p(d|u_d, l_d) \cdot \prod_{d \in \mathcal{D}} \prod_{t\in \mathcal{T}_d} p(t|d) \cdot \\
& \prod_{d\in \mathcal{D}}\prod_{w_i} p(w_i|d) p(\mathcal{C}(w_i, h)|w_i),
\label{eqn:obj}
\end{split}
\end{equation}
where $u_d$ is the user creating document $d$; $l_d$ is the label of document $d$; $\mathcal{T}_d$ is the set of tags in $d$. If $d$ is not labeled (i.e., $l_d$ is unknown), according to Eq. (\ref{eqn:lx}),
\begin{equation}
p(d|u_d, l_d) = \frac{\exp(\bme_d^T\bme_{u_d})}{\sum_{d'\in \mathcal{D}}\exp(\bme_{d'}^T\bme_{u_d})}.
\end{equation}
If $d$ is labeled, according to Eq. (\ref{eqn:ruf}),
\begin{equation}
\begin{split}
p(d|u_d, l_d) &= \frac{\exp(\bme_d^T\bme_{u_d})\exp(\bme_d^T\bme_{l_d})}{\sum_{d'\in \mathcal{D}}\exp(\bme_{d'}^T\bme_{u_d})\exp(\bme_{d'}^T\bme_{l_d})} \\
&\propto \frac{\exp(\bme_d^T\bme_{u_d})}{\sum_{d'\in \mathcal{D}}\exp(\bme_{d'}^T\bme_{u_d})}\cdot \frac{\exp(\bme_d^T\bme_{l_d})}{\sum_{d'\in \mathcal{D}}\exp(\bme_{d'}^T\bme_{l_d})}.
\end{split}
\end{equation}
Similarly, other conditional probabilities can be derived using Eqs. (\ref{eqn:wr})-(\ref{eqn:wcf}). Therefore,
\begin{equation}
\small
\begin{split}
\log\mathcal{J} =&\sum_{u\in \mathcal{U}}\sum_{d\in \mathcal{D}_u}\log \frac{\exp(\bme_d^T\bme_u)}{\sum_{d'}\exp(\bme_{d'}^T\bme_u)} 
+ \sum_{l\in \mathcal{L}}\sum_{d\in \mathcal{D}_l}\log \frac{\exp(\bme_d^T\bme_l)}{\sum_{d'}\exp(\bme_{d'}^T\bme_l)} \\
+ & \sum_{d \in \mathcal{D}} \sum_{t\in \mathcal{T}_d}\log \frac{\exp(\bme_t^T\bme_d)}{\sum_{t'}\exp(\bme_{t'}^T\bme_d)}
+ \sum_{d\in \mathcal{D}}\sum_{w_i}\log \frac{\exp(\bme_{w_i}^T\bme_d)}{\sum_{w'}\exp(\bme_{w'}^T\bme_d)} \\
+ & \sum_{d\in \mathcal{D}}\sum_{w_i}\sum_{w_j \in \mathcal{C}(w_i, h)} \log \frac{\exp(\bme'^{\ T}_{w_j} \bme_{w_i})}{\sum_{w'}\exp(\bme'^{\ T}_{w'} \bme_{w_i})} + \text{const}. 
\end{split}
\label{eqn:logl}
\end{equation}
Here $\mathcal{U}$ is the set of users in the dataset; $\mathcal{D}_u$ is the set of documents belonging to user $u$; $\mathcal{D}_l$ is the set of documents (in the training set) with label $l$.
$\bme_u$, $\bme_l$, $\bme_d$, $\bme_t$, $\bme_{w}$ and $\bme'_{w}$ can be learned by maximizing 
$\log\mathcal{J}$. However, the denominators in Eq. (\ref{eqn:logl}) require summing over all documents/tags/words, which is computationally expensive. Following common practice, in our actual computation, we estimate these
terms through negative sampling \cite{mikolov2013distributed,tang2015line}.


\vspace{1mm}

\noindent \textbf{Comparisons with Heterogeneous Network Embedding.} There are various ways to encode multi-modal signals, among which Heterogeneous Information Network (HIN) embedding \cite{shang2016meta,dong2017metapath2vec,fu2017hin2vec} is commonly used. In fact, if we remove the edge directions in Figure \ref{fig:story}(a), it can also be viewed as an HIN schema \cite{sun2011pathsim}. We would like to emphasize two key differences between our method and HIN embedding. First, HIN models \textit{connections} between different types of nodes, while our approach models \textit{generative relationships}. Not all connections can be explained as a generative story. From this perspective, our method is more specifically designed to encode text with metadata information. Second, many HIN embedding methods \cite{shang2016meta,dong2017metapath2vec} require users to specify a set of meta-paths \cite{sun2011pathsim}, which is not needed in our approach. 

\vspace{-0.9mm}
\subsection{Training Data Generation}
\label{sec:pseudo}
To deal with label scarcity, we consider to generate synthesized training data. 
In Figure \ref{fig:story}, we have proposed a generative process to characterize heterogeneous signals in both text and metadata. Our embedding module follows this process, and so will our generation module. To be specific, given a label $l$, we first generate a document vector $e_d$ from $p(d|u, l)$ and then sample words and tags one by one from $p(w|d)$ and $p(t|d)$. 
Formally, 
\begin{equation}
    p(d,w_{1:n},t_{1:m}|u,l) = p(d|u,l) \cdot  \prod_{i=1}^n p(w_i|d) \cdot \prod_{j=1}^m p(t_j|d).
\label{eqn:gen}
\end{equation}
Unlike in the case of embedding, when we tend to generate new documents, $u$ is not observable. Also, there is no need to generate training data for a specific existing user because our task is to predict the label, not the user, of each document. If $u$ is assumed to be unknown, following the derivation of Eqs. (\ref{eqn:nol})-(\ref{eqn:lx}), we have
\begin{equation}
    p(d|u,l) \propto \exp(\bme_d^T\bme_l).
\end{equation}
In principle, we are able to generate as many documents as we want. In other words, there is an infinite number of documents (i.e., $\bme_d$) distributed on the sphere, and we are essentially picking some of them as synthesized training samples. Similar to Eq. (\ref{eqn:vmf0}), by assuming $|\mathcal{D}|\rightarrow\infty$,  
we have
\begin{equation}
p(d|u,l) = \text{vMF}_p(\bme_l, \kappa).
\label{eqn:rl2}
\end{equation}

The generation of words and tags, however, is different. The semantics of words and tags are discretely distributed in the latent space. To be specific, not every unit vector can be mapped back to an existing word or tag in the vocabulary. Therefore, we still assume $|\mathcal{V}|$ and $|\mathcal{T}|$ are finite in our generation step, and the generated words or tags must have appeared in $\mathcal{V}$ or $\mathcal{T}$. Then the generation probabilities simply follow Eqs. (\ref{eqn:wr}) and (\ref{eqn:tr}). In practice, the computation of $\sum_{w'\in \mathcal{V}}\exp(\bme_{w'}^T\bme_d)$ could be quite expensive. Therefore, we restrict the word/tag candidates to be the top-$\tau$ ones similar with $\bme_d$ on the sphere (denoted as $\mathcal{N}(\bme_d)$), and the conditional probabilities will be modified as

\begin{equation}
p(w_i|d) = \frac{\exp(\bme_{w_i}^T\bme_d)}{\sum_{w'\in \mathcal{N}(\bme_d)}\exp(\bme_{w'}^T\bme_d)},\ w_i \in \mathcal{N}(\bme_d),
\label{eqn:wr2}
\end{equation}
and
\begin{equation}
p(t_j|d) = \frac{\exp(\bme_{t_j}^T\bme_d)}{\sum_{t'\in \mathcal{N}(\bme_d)}\exp(\bme_{t'}^T\bme_d)},\ \  t_j \in \mathcal{N}(\bme_d).
\label{eqn:wr3}
\end{equation}

Plugging Eqs. (\ref{eqn:rl2}), (\ref{eqn:wr2}) and (\ref{eqn:wr3}) into Eq. (\ref{eqn:gen}) fully specifies our generation step. Each generated document has a label, a sequence of words and a set of tags. We denote the set of synthesized training document for class $l$ as $\mathcal{D}_l^*$.

\subsection{Neural Model Training}
\label{sec:neural}
We now need to feed the synthesized training data $\{\mathcal{D}_l^*:l\in \mathcal{L}\}$, together with the ``real'' training data $\{\mathcal{D}_l:l\in \mathcal{L}\}$, into a classifier. Here each document (either synthesized or real) can be viewed as a sequence of words $w_{1:n}$ and tags $t_{1:m}$. The input of the classifier is pre-trained embeddings $\{\bme_w:w\in \mathcal{V}\}$ and $\{\bme_t:t\in \mathcal{T}\}$. This setting is generic enough so that many neural classifiers, such as CNN \cite{kim2014convolutional}, HAN \cite{yang2016hierarchical}, CNN-LSTM \cite{zhou2015c}, BiGRU-CNN \cite{wang2017hybrid} and DPCNN \cite{johnson2017deep}, can be applied in our framework. Developing an advanced neural classifier is not a goal of this paper. Instead, we aim to show the power of our embedding and generation modules. Their contribution to the performance should not be covered by advanced neural architectures when comparing with baselines. Therefore, we choose CNN \cite{kim2014convolutional}, a simple but widely used model as our text classifier. We also tried HAN \cite{yang2016hierarchical}, a representative RNN-based classifier, but it performs slightly worse than CNN on our datasets.

In the CNN architecture, each document is represented by the concatenation of its word and tag embeddings $\bme_{1:n+m} = [\bme_{w_1},...,$ $\bme_{w_n}, \bme_{t_1},...,\bme_{t_m}]$. There is one convolutional layer, in which a convolution filter is applied to a text region $\bme_{i:i+h-1}$. 
\begin{equation}
c_i = \sigma({\bm w}^T\bme_{i:i+h-1}+{\bm b}),
\end{equation}
where $\sigma$ is the sigmoid function. All $c_i$'s together form a feature map ${\bm c} = [c_1,...,c_{n+m-h+1}]$ associated with the filter. Then a max-over-time pooling operation is performed on ${\bm c}$. In this paper, we use filters with $h = 2, 3, 4$ and 5. For each width $h$, we generate 20 feature maps. After pooling, the features are passed through a fully connected softmax layer whose output is the probability distribution over labels.

\newcommand{\bml}{{\bm l}}
\newcommand{\bmq}{{\bm q}}

Following \cite{yang2016hierarchical}, we use negative log-likelihood of the correct labels as training loss. Formally, given all training samples (including both real and synthesized ones), assume the $i$-th one has label $lb(i)$. Then
\begin{equation}
\textit{loss} = -\sum_{i}\log q_{i,lb(i)},
\end{equation}
where $\bmq_i$ is the neural network output distribution of the $i$-th instance. 

\subsection{The General Version of Our Framework}
\label{sec:general}
Although we illustrate our framework using Figure \ref{fig:story}(a), \textsc{MetaCat} can be easily generalized to any generative process following Figure \ref{fig:story}(b). To be specific, let $\mathcal{M}_{\mathcal{G}}$ be the set of global metadata variables and $\mathcal{M}_{\mathcal{L}}$ be the set of local metadata variables. (For example, in the cases of GitHub repositories and tweets, $\mathcal{M}_{\mathcal{G}} = \{user\}$ and $\mathcal{M}_{\mathcal{L}} = \{tag\}$; in the case of Amazon reviews, $\mathcal{M}_{\mathcal{G}} = \{user, product\}$ and $\mathcal{M}_{\mathcal{L}} = \emptyset$.) The generative process can be described as follows:

\vspace{1mm}

\noindent \textbf{Global Metadata \& Label $\rightarrow$ Document.}
\begin{equation}
    p(d|\mathcal{M}_{\mathcal{G}}, l) \propto \exp(\bme_d^T\bme_l)\cdot \prod_{z \in \mathcal{M}_{\mathcal{G}}} \exp(\bme_d^T\bme_z).
\end{equation}

\noindent \textbf{Document $\rightarrow$ Word \& Local Metadata.}
\begin{equation}
    p(w, \mathcal{M}_{\mathcal{L}}|d) \propto \exp(\bme_w^T\bme_d)\cdot \prod_{z \in \mathcal{M}_{\mathcal{L}}} \exp(\bme_z^T\bme_d).
\end{equation}

\noindent \textbf{Word $\rightarrow$ Context.} 
\begin{equation}
p(\mathcal{C}(w_i, h)|w_i) \propto \prod_{w_j \in \mathcal{C}(w_i, h)}\exp(\bme'^{\ T}_{w_j} \bme_{w_i}).
\end{equation}

Given the generative process, the derivations of embedding learning and training data generation are quite similar to those in Sections \ref{sec:embedding} and \ref{sec:pseudo}. We directly show the results here.

\vspace{1mm}

\noindent \textbf{Embedding Learning.} The log-likelihood is 
\begin{equation}
\small
\begin{split}
\log\mathcal{J} =& \sum_{z \in \mathcal{Z}_{\mathcal{G}}}\sum_{d\in \mathcal{D}_z}\log \frac{\exp(\bme_d^T\bme_z)}{\sum_{d'}\exp(\bme_{d'}^T\bme_z)} + \sum_{l\in \mathcal{L}}\sum_{d\in \mathcal{D}_l}\log \frac{\exp(\bme_d^T\bme_l)}{\sum_{d'}\exp(\bme_{d'}^T\bme_l)} \\
+ & \sum_{d \in \mathcal{D}} \sum_{z\in \mathcal{Z}_d}\log \frac{\exp(\bme_z^T\bme_d)}{\sum_{z'}\exp(\bme_{z'}^T\bme_d)} +  \sum_{d\in \mathcal{D}}\sum_{w_i}\log \frac{\exp(\bme_{w_i}^T\bme_d)}{\sum_{w'}\exp(\bme_{w'}^T\bme_d)} \\
+ & \sum_{d\in \mathcal{D}}\sum_{w_i}\sum_{w_j \in \mathcal{C}(w_i, h)} \log \frac{\exp(\bme'^{\ T}_{w_j} \bme_{w_i})}{\sum_{w'}\exp(\bme'^{\ T}_{w'} \bme_{w_i})} + \text{const}. 
\end{split}
\end{equation}
Here $\mathcal{Z}_{\mathcal{G}}$ is the set of global metadata instances (e.g., $\mathcal{Z}_{\mathcal{G}} = \mathcal{U}$ for GitHub repositories and tweets, and $\mathcal{Z}_{\mathcal{G}} = \mathcal{U}\cup\mathcal{P}$ for Amazon reviews, where $\mathcal{P}$ is the set of products); $\mathcal{D}_z$ is the set of documents belonging to a global metadata instance $z$; $\mathcal{Z}_d$ is the set of local metadata instances associated with document $d$ (e.g., $\mathcal{Z}_d = \mathcal{T}_d$ for GitHub repositories and tweets, and $\mathcal{Z}_d = \emptyset$ for Amazon reviews). All embeddings can be learned by maximizing this likelihood.

\vspace{1mm}

\noindent \textbf{Training Data Generation.} Given label $l$, we first generate $e_d$ from $p(d|\mathcal{M}_{\mathcal{G}}, l)$ and then sample words $w_i$ and local metadata instances $z_j$ from $p(w|d)$ and $p(z|d)$, respectively.
\begin{equation}
p(d,w_{1:n},z_{1:m}|\mathcal{M}_{\mathcal{G}},l) = p(d|\mathcal{M}_{\mathcal{G}},l) \cdot  \prod_{i=1}^n p(w_i|d) \cdot \prod_{j=1}^m p(z_j|d).
\label{eqn:gen1}
\end{equation}
The generation of documents follows Eq. (\ref{eqn:rl2}). By assuming $\mathcal{M}_{\mathcal{G}}$ is not observable and $|\mathcal{D}|\rightarrow\infty$, we have
\begin{equation}
p(d|\mathcal{M}_{\mathcal{G}},l) = \text{vMF}_p(\bme_l, \kappa).
\end{equation}
The generation of words and local metadata follows Eqs. (\ref{eqn:wr2}) and (\ref{eqn:wr3}), where we only need to replace $t_j$ with $z_j$.

\section{Experiments}
In order to provide evidence for the efficacy of \textsc{MetaCat}, we experiment with five datasets collected from different domains. Specifically, we aim to show that (1) \textsc{MetaCat} outperforms benchmark approaches by a clear margin; (2) the proposed generation-guided embedding module captures multi-modal semantics better than existing meta-path based HIN embedding techniques; (3) incorporating synthesized training samples can significantly boost the classification performance.

\subsection{Datasets} 
The five datasets we use are collected from three different sources: GitHub repositories, Amazon reviews and Twitter posts.\footnote{Our code and datasets are available at \\ \texttt{\url{https://github.com/yuzhimanhua/MetaCat}}.}
\begin{itemize}[leftmargin=*]
\item \textbf{GitHub-Bio \cite{zhang2019higitclass}.} This dataset is extracted from four bioinformatics venues from 2014 to 2018. Each GitHub repository is associated with a research article published on these venues. The issue section (e.g., sequence analysis, genome analysis, systems biology, etc.) of the article is viewed as the topic label of the associated repository.

\item \textbf{GitHub-AI \cite{zhang2019higitclass}.} This dataset is collected by the Paper With Code project.
It contains a list of GitHub repositories implementing algorithms of various machine learning tasks (e.g., image generation, machine translation, speech recognition, etc.). Each task is viewed as a topic label.
 
\item \textbf{GitHub-Sec.} This dataset is obtained from the DARPA SocialSim Challenge. It contains GitHub repositories related to cryptocurrency, cybersecurity or software vulnerability.

\item \textbf{Amazon \cite{mcauley2013hidden}.}  This dataset is a large crawl of Amazon product reviews. The topic label of each review is its product category (e.g., books, home \& kitchen, sports \& outdoors, etc.). We select 10 large categories and sample 10,000 reviews from each category.

\item \textbf{Twitter \cite{zhang2017react}.} This dataset contains geo-tagged tweets in New York City during 2014.08.01 -- 2014.11.30. The authors link the tweets with Foursquare's POI database, and the topic label of each tweet is the type of the linked POI (e.g., shop \& service, college \& university, nightlife spot, etc.).
\end{itemize}

We use 10 documents in each class for training and all the others for testing (i.e., $|\mathcal{D}_l| = 10$). Brief statistics of the five datasets are summarized in Table \ref{tab:data}.

\begin{table}[!t]
	\caption{Dataset Statistics.}
	\vspace{-1em}
	\scalebox{0.92}{
		\begin{tabular}{c|cccc}
			\hline
			Dataset          & \#Document   & \#Class  & \#Training & \#Testing \\
			\hline
			GitHub-Bio \cite{zhang2019higitclass}      & 876      & 10       & 100     & 776    \\
			GitHub-AI \cite{zhang2019higitclass}        & 1,596    & 14       & 140     & 1,456     \\
			GitHub-Sec       & 84,950   & 3        & 30      & 84,920  \\
			Amazon \cite{mcauley2013hidden}          & 100,000  & 10       & 100     & 99,900  \\
			Twitter \cite{zhang2017react}          & 135,529  & 9        & 90      & 135,439  \\
			\hline
		\end{tabular}
	}
	\vspace{-1em}
	\label{tab:data}
\end{table}

\begin{table*}[]
\centering
\caption{Micro F1 scores of compared algorithms on the five datasets. ``--'': excessive memory requirements.}
\vspace{-0.5em}
\scalebox{0.92}{
\begin{tabular}{c|c|c|c|c|c|c}
	\hline
	Type                         & Method       & GitHub-Bio    & GitHub-AI     & GitHub-Sec    & Amazon        & Twitter       \\ \hline
	\multirow{6}{*}{Text-based}  & CNN \cite{kim2014convolutional}         & 0.2227 $\pm$ 0.0195 & 0.2404 $\pm$ 0.0404 & 0.4909 $\pm$ 0.0489 & 0.4915 $\pm$ 0.0374 & 0.3106 $\pm$ 0.0613 \\
	& HAN \cite{yang2016hierarchical}         & 0.1409 $\pm$ 0.0145 & 0.1900 $\pm$ 0.0299 & 0.4677 $\pm$ 0.0334 & 0.4809 $\pm$ 0.0372 & 0.3163 $\pm$ 0.0878 \\
	& PTE \cite{tang2015pte}          & 0.3170 $\pm$ 0.0516 & 0.3511 $\pm$ 0.0403 & 0.4551 $\pm$ 0.0249 & 0.2997 $\pm$ 0.0786 & 0.1945 $\pm$ 0.0250 \\
	& WeSTClass \cite{meng2018weakly}    & 0.3680 $\pm$ 0.0138 & 0.5036  $\pm$ 0.0287 & 0.6146 $\pm$ 0.0084 & 0.5312 $\pm$ 0.0161 & 0.3568 $\pm$ 0.0178 \\
	& PCEM \cite{xiao2019efficient}         & 0.3426 $\pm$ 0.0160       & 0.4820 $\pm$ 0.0292       & 0.5912 $\pm$ 0.0341       & 0.4645 $\pm$ 0.0163        &  0.2387 $\pm$ 0.0344             \\
	& BERT \cite{devlin2019bert}        &   0.2680 $\pm$ 0.0303           &  0.2451 $\pm$ 0.0273             &  0.5538 $\pm$ 0.0368             &  0.5240 $\pm$ 0.0261              &   0.3312 $\pm$ 0.0860            \\ \hline
	\multirow{4}{*}{Graph-based} & ESim \cite{shang2016meta}         & 0.2925 $\pm$ 0.0223 & 0.4376 $\pm$ 0.0323 & 0.5480 $\pm$ 0.0109 & 0.5320 $\pm$ 0.0246 & 0.3512 $\pm$ 0.0226 \\
	& Metapath2vec \cite{dong2017metapath2vec} & 0.3956 $\pm$ 0.0141 & 0.4444 $\pm$ 0.0231 & 0.5772 $\pm$ 0.0594 & 0.5256 $\pm$ 0.0335 & 0.3516 $\pm$ 0.0407 \\
	& HIN2vec \cite{fu2017hin2vec}      & 0.2564 $\pm$ 0.0131 & 0.3614 $\pm$ 0.0234 & 0.5218 $\pm$ 0.0466 & 0.4987 $\pm$ 0.0252 & 0.2944 $\pm$ 0.0614              \\
	& TextGCN \cite{yao2019graph}      & 0.4759 $\pm$ 0.0126 & 0.6353 $\pm$ 0.0059 & --             & --             & 0.3361 $\pm$ 0.0032 \\ \hline
	& \textsc{MetaCat}      & \textbf{0.5258 $\pm$ 0.0090} & \textbf{0.6889 $\pm$ 0.0128} & \textbf{0.7243 $\pm$ 0.0336} & \textbf{0.6422 $\pm$ 0.0058} & \textbf{0.3971 $\pm$ 0.0169} \\ \hline
\end{tabular}
}
\vspace{-0.5em}
\label{tab:acc}
\end{table*}

\begin{table*}[]
\centering
\caption{Macro F1 scores of compared algorithms on the five datasets. ``--'': excessive memory requirements.}
\vspace{-0.5em}
\scalebox{0.92}{
	\begin{tabular}{c|c|c|c|c|c|c}
		\hline
		Type                         & Method       & GitHub-Bio    & GitHub-AI     & GitHub-Sec    & Amazon        & Twitter       \\ \hline
		\multirow{6}{*}{Text-based}  & CNN \cite{kim2014convolutional}         & 0.1896 $\pm$ 0.0133 & 0.1796 $\pm$ 0.0216 & 0.4268 $\pm$ 0.0584 & 0.5056 $\pm$ 0.0376 & 0.2858 $\pm$ 0.0559 \\
		& HAN \cite{yang2016hierarchical}         & 0.0677 $\pm$ 0.0208 & 0.0961 $\pm$ 0.0254 & 0.4095 $\pm$ 0.0590 & 0.4644 $\pm$ 0.0597 & 0.2592 $\pm$ 0.0826 \\
		& PTE \cite{tang2015pte}          & 0.2630 $\pm$ 0.0371 & 0.3363 $\pm$ 0.0250 & 0.3803 $\pm$ 0.0218 & 0.2563 $\pm$ 0.0810 & 0.1739 $\pm$ 0.0190 \\
		& WeSTClass \cite{meng2018weakly}    & 0.3414 $\pm$ 0.0129 & 0.4056 $\pm$ 0.0248 & 0.5497 $\pm$ 0.0054 & 0.5234 $\pm$ 0.0147 & 0.3085 $\pm$ 0.0398 \\
		& PCEM \cite{xiao2019efficient}         & 0.2977 $\pm$ 0.0281       & 0.3751 $\pm$ 0.0350       & 0.4033 $\pm$ 0.0336       & 0.4239 $\pm$ 0.0237       & 0.2039       $\pm$ 0.0472       \\
		& BERT \cite{devlin2019bert}         &  0.1740 $\pm$ 0.0164             & 0.2083 $\pm$ 0.0415              &  0.4956 $\pm$ 0.0164             &  0.4911 $\pm$ 0.0544             & 0.2834 $\pm$ 0.0550            \\ \hline
		\multirow{4}{*}{Graph-based} & ESim \cite{shang2016meta}         & 0.2598 $\pm$ 0.0182 & 0.3209 $\pm$ 0.0202 & 0.4672 $\pm$ 0.0171 & 0.5336 $\pm$ 0.0220 & 0.3399 $\pm$ 0.0113 \\
		& Metapath2vec \cite{dong2017metapath2vec} & 0.3214 $\pm$ 0.0128 & 0.3220 $\pm$ 0.0290 & 0.5140 $\pm$ 0.0637 & 0.5239 $\pm$ 0.0437 & 0.3443 $\pm$ 0.0208 \\
		& HIN2vec \cite{fu2017hin2vec}      & 0.2742 $\pm$ 0.0136 & 0.2513 $\pm$ 0.0211 & 0.4000 $\pm$ 0.0115 & 0.4261 $\pm$ 0.0284 & 0.2411 $\pm$ 0.0142              \\
		& TextGCN \cite{yao2019graph}      & 0.4817 $\pm$ 0.0078 & 0.5997 $\pm$ 0.0013 & --             & --             & 0.3191 $\pm$ 0.0029 \\ \hline
		& \textsc{MetaCat}      & \textbf{0.5230 $\pm$ 0.0080} & \textbf{0.6154 $\pm$ 0.0079} & \textbf{0.6323 $\pm$ 0.0235} & \textbf{0.6496 $\pm$ 0.0091} & \textbf{0.3612 $\pm$ 0.0067} \\ \hline
	\end{tabular}
}
\vspace{-0.5em}
\label{tab:accmacro}
\end{table*}

\vspace{1mm}

\subsection{Baseline Methods}
We evaluate the performance of \textsc{MetaCat} against both text-based and graph-based benchmark approaches: 
\begin{itemize}[leftmargin=*]
\item \textbf{CNN \cite{kim2014convolutional}} is a supervised text classification method. It trains a convolutional neural network with a max-over-time pooling layer. 

\item \textbf{HAN \cite{yang2016hierarchical}} is a supervised text classification method. It trains a hierarchical attention network and uses GRU to encode word sequences. 

\item \textbf{PTE \cite{tang2015pte}} is a semi-supervised approach. It constructs a network with three subgraphs (word-word, word-document and word-label) and embeds nodes based on first and second order proximities.

\item \textbf{WeSTClass \cite{meng2018weakly}} is a weakly supervised text classification approach. It models topic semantics in the word2vec embedding space and applies a pre-training and self-training scheme.

\item \textbf{PCEM \cite{xiao2019efficient}} is a weakly supervised hierarchical text classification method using path cost-sensitive learning. In our problem setting, the label hierarchy has only one layer.

\item \textbf{BERT \cite{devlin2019bert}} is a state-of-the-art pre-trained language model that provides contextualized word representations. Here we fine-tune BERT under the supervised text classification setting using labeled documents.

\item \textbf{ESim \cite{shang2016meta}} is an HIN embedding approach. It learns node embeddings using meta-path guided sequence sampling and noise-contrastive estimation.

\item \textbf{Metapath2vec \cite{dong2017metapath2vec}} is an HIN embedding approach. It samples node sequences through heterogeneous random walks and incorporates negative sampling.  

\item \textbf{HIN2vec \cite{fu2017hin2vec}} is an HIN embedding approach that exploits different types of links among nodes.

\item \textbf{TextGCN \cite{yao2019graph}} is a semi-supervised text classification approach. It applies graph neural networks on the document-word co-occurrence graph.
\end{itemize}

For ESim, Metapath2vec and HIN2vec, we construct an HIN by viewing Figure \ref{fig:story} as the HIN schema. Based on the schema, we select five meta-paths $doc$-$user$-$doc$, $doc$-$label$-$doc$, $doc$-$word$-$doc$, $doc$-$tag$-$doc$ and $word$-$word$ to guide the embedding. (On Amazon, we replace $doc$-$tag$-$doc$ with $doc$-$product$-$doc$.)
After the node embedding step, we get representation vectors of each word/tag. Then, we train a CNN classifier using original training data $\{\mathcal{D}_l:l\in \mathcal{L}\}$ and the pre-trained HIN embeddings. 

For all baselines using the CNN classifier (i.e., CNN, WeSTClass, ESim, Metapath2vec and HIN2vec), we adopt the architecture in Section \ref{sec:neural} (the same as \textsc{MetaCat}) to align the experiment settings. For HAN, we use a forward GRU with 100 dimension for both word and sentence encoding. The training process of all neural classifiers is performed using SGD with a batch size of 256. The dimension of all embedding vectors is 100 (except BERT whose base model is set to be 768-dimensional).

For \textsc{MetaCat}, we set the local context window size $h=5$, the document-specific vocabulary size $|\mathcal{N}(\bme_d)|=50$, and the number of generated training samples per class $|\mathcal{D}_l^*| = 100$.

\subsection{Performance Comparison}
Tables \ref{tab:acc} and \ref{tab:accmacro} demonstrate the Micro and Macro F1 scores of compared methods on the five datasets. We repeat each experiment 5 times with the mean and standard deviation reported. We cannot get the performance of TextGCN on GitHub-Sec and Amazon because the model (i.e., the constructed graph) is too large to fit into our GPU with 11GB memory. (The Twitter dataset can fit because it has a much smaller average document length.)

As we can observe from Tables \ref{tab:acc} and \ref{tab:accmacro}: (1) \textsc{MetaCat} consistently outperforms all baselines by a clear margin on all datasets. It achieves a 4.3\% \textit{absolute} improvement on average in comparison with TextGCN, the second best approach in our table. When comparing with other baselines, our absolute improvement is over 10\%. 
(2) In contrast to methods using plain text embeddings (i.e., CNN, HAN and WeSTClass), the performance boosts of \textsc{MetaCat} are more significant on smaller datasets (i.e., GitHub-Bio and GitHub-AI). In fact, when the corpus is small, word2vec cannot generate high-quality word embeddings. Consequently, neural models using word2vec embeddings will not achieve satisfying performance. In this case, leveraging multi-modal signals in representation learning becomes necessary. 
(3) Despite its great success in supervised tasks, BERT is not suitable for our task without sufficient training data, probably because the language style of GitHub files and tweets are different from that of Wikipedia, which might require strong supervision for fine-tuning.
(4) Although HIN embedding techniques (i.e., ESim, Metapath2vec and HIN2vec) consider data heterogeneity and utilize unlabeled data during the embedding step, they still do not address the label scarcity bottleneck when training the classifier. This leads to their inferiority towards \textsc{MetaCat}.

\subsection{Effect of Embedding Learning}
\label{sec:hin}
\noindent \textbf{Embedding Method.} Tables \ref{tab:acc} and \ref{tab:accmacro} tell us that our generation-guided embedding module, \textit{together with training data generation}, can outperform several HIN embedding baselines. Now, to further explore the effectiveness of the proposed embedding technique, we perform a ``fairer'' comparison by fixing all the other parts in \textsc{MetaCat} and vary the embedding module only. 

\begin{figure*}[t]
\centering
\subfigure[\textsc{GitHub-Bio}]{
\includegraphics[height=2.8cm]{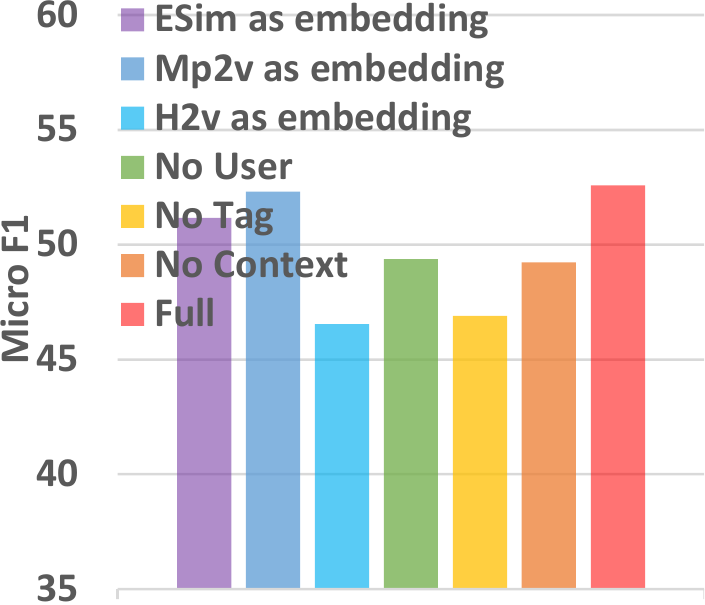}}
\subfigure[\textsc{GitHub-AI}]{
\includegraphics[height=2.8cm]{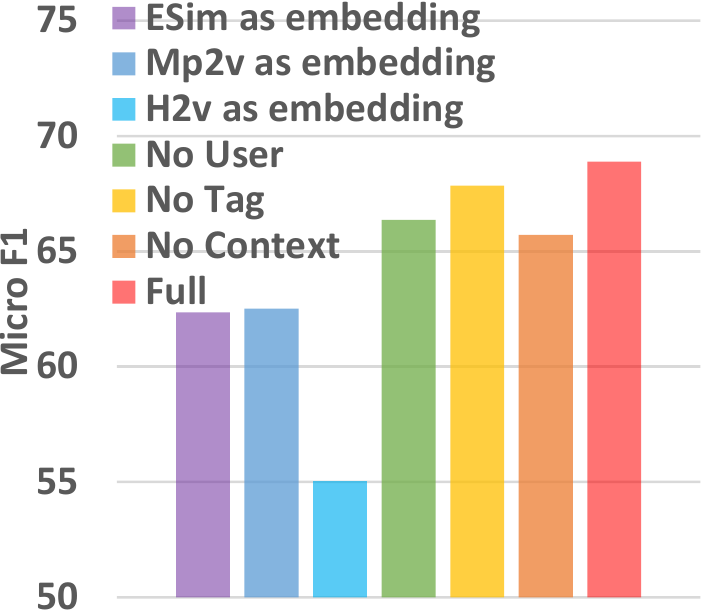}}
\subfigure[\textsc{GitHub-Sec}]{
\includegraphics[height=2.8cm]{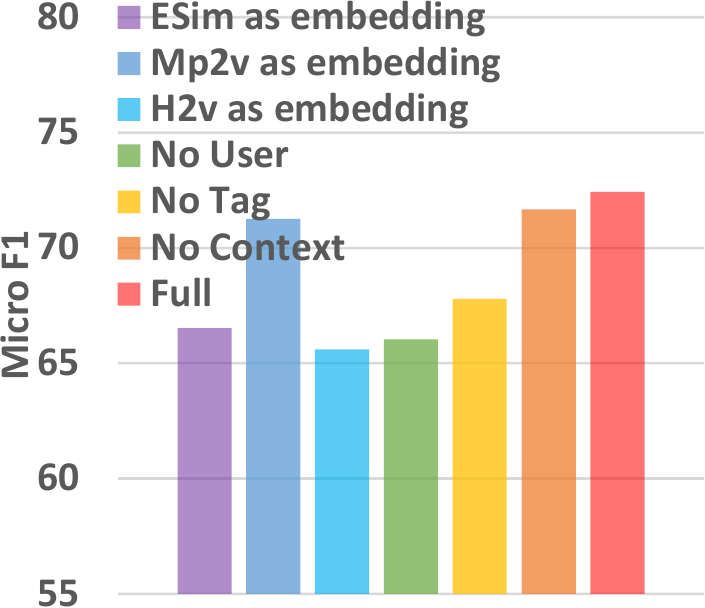}}
\subfigure[\textsc{Amazon}]{
\includegraphics[height=2.8cm]{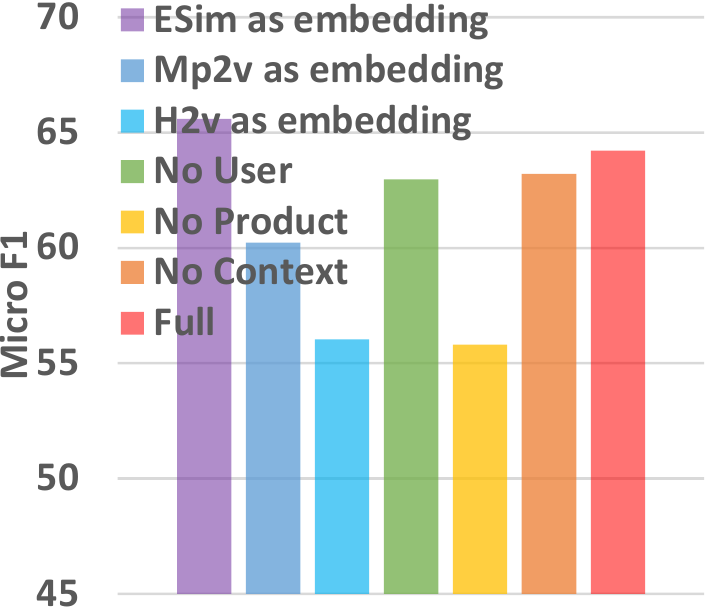}}
\subfigure[\textsc{Twitter}]{
\includegraphics[height=2.8cm]{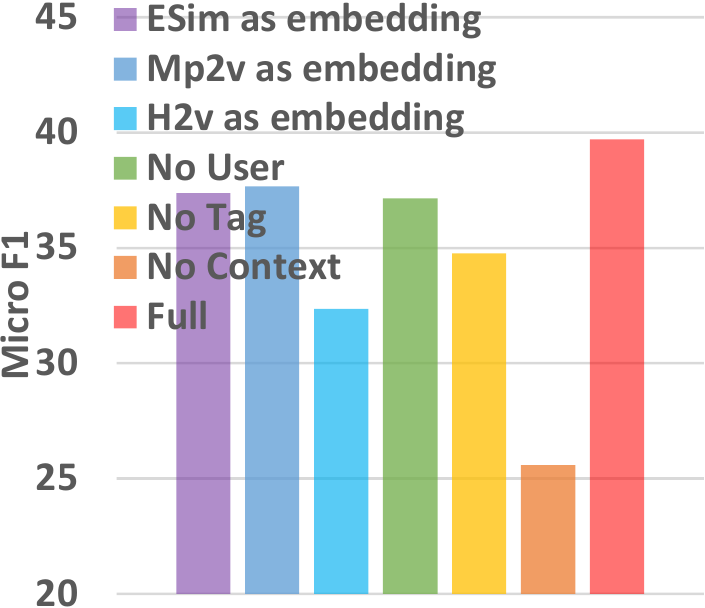}}
\vspace{-1em}
\caption{Micro F1 scores of algorithms with different embedding modules.} 
\vspace{-1em}
\label{fig:embabl}
\end{figure*}

\begin{figure*}[t]
	\centering
	\subfigure[\textsc{GitHub-Bio}]{
		\includegraphics[height=2.8cm]{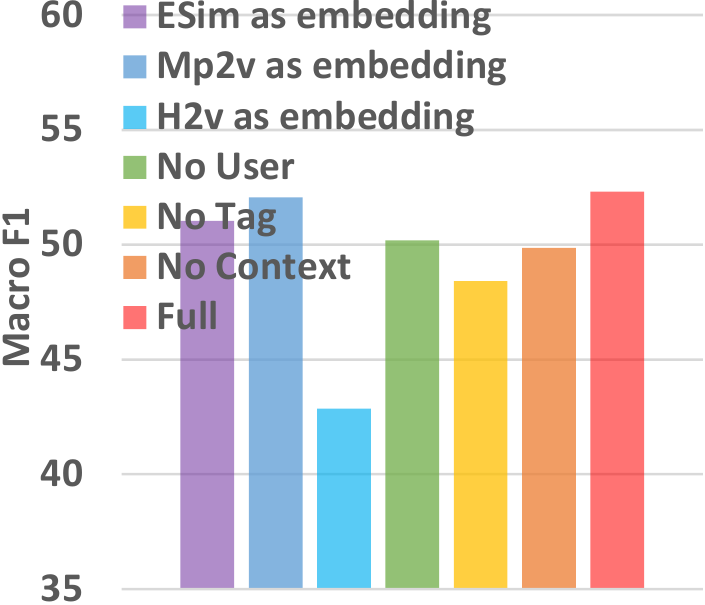}}
	\subfigure[\textsc{GitHub-AI}]{
		\includegraphics[height=2.8cm]{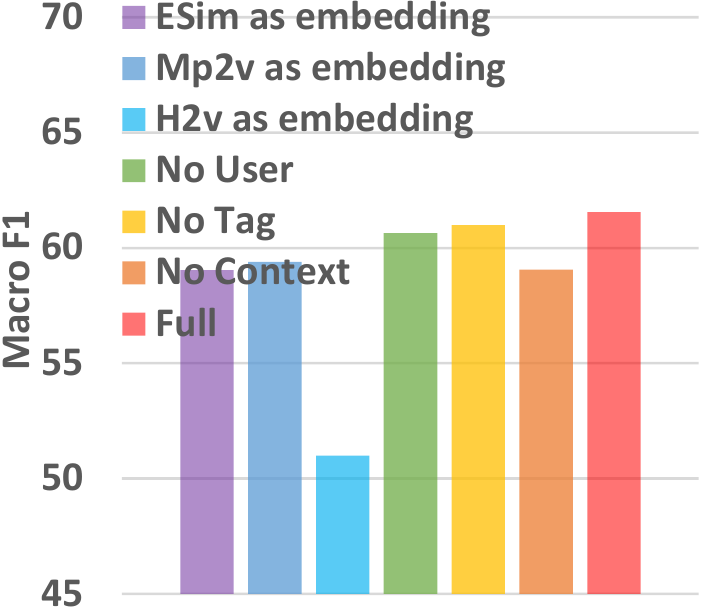}}
	\subfigure[\textsc{GitHub-Sec}]{
		\includegraphics[height=2.8cm]{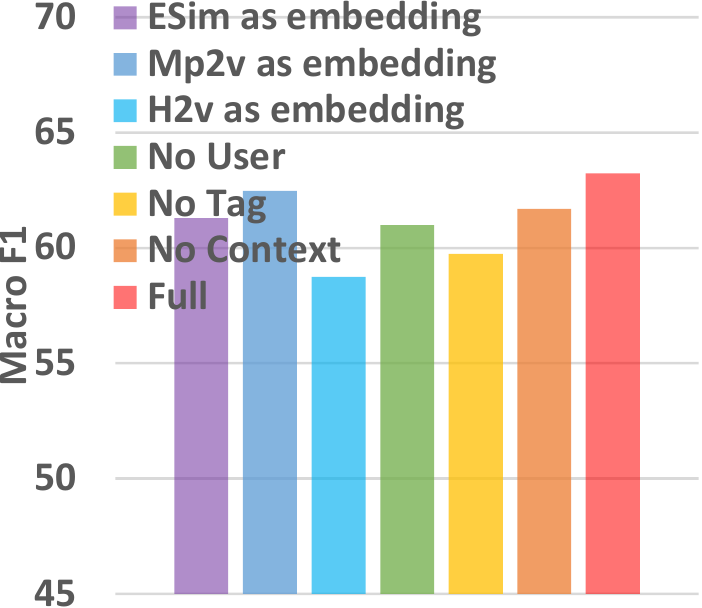}}
	\subfigure[\textsc{Amazon}]{
		\includegraphics[height=2.8cm]{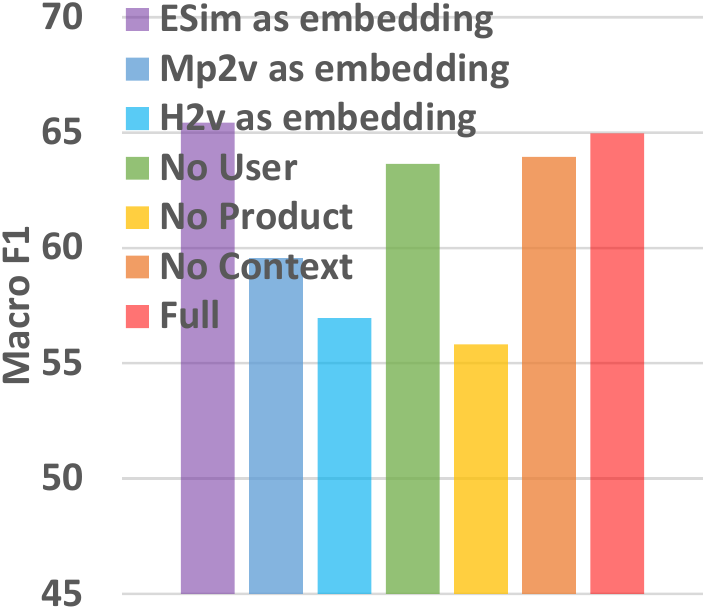}}
	\subfigure[\textsc{Twitter}]{
		\includegraphics[height=2.8cm]{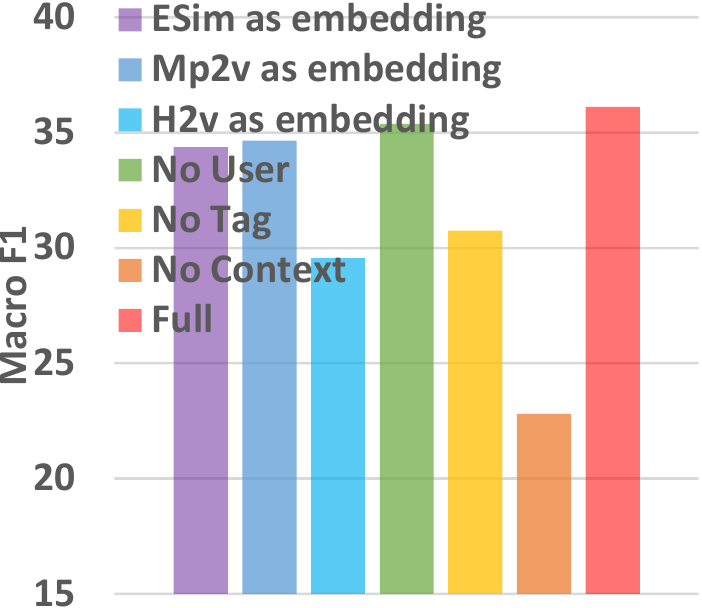}}
	\vspace{-1em}
	\caption{Macro F1 scores of algorithms with different embedding modules.} 
	\vspace{-1em}
	\label{fig:embablmacro}
\end{figure*}

To be specific, we can use ESim, Metapath2vec and HIN2vec to replace our current embedding technique, which generates three ablations \textsc{ESim-as-embedding}, \textsc{Mp2v-as-embedding} and \textsc{H2v-as-embedding}. Moreover, we exploit various metadata as well as word context information during the embedding process. To demonstrate their contributions, we create three ablations \textsc{No-User}, \textsc{No-Context}, and \textsc{No-Tag} (\textsc{No-Product} for the Amazon dataset). Here \textsc{No-User} means we do not consider user information during embedding. Similar meanings can be inferred for the other two ablations. Note that our generation step needs $\bme_l$ and $\bme_w$, so there is no so-called ``\textsc{No-Label}'' or ``\textsc{No-Word}''. Figures \ref{fig:embabl} and \ref{fig:embablmacro} show the performance of these variants and our \textsc{Full} model.

We have the following observations: (1) \textsc{Full} outperforms \textsc{ESim-as-embedding}, \textsc{Mp2v-as-embedding} and \textsc{H2v-as-embedding} on almost all datasets, with only one exception where \textsc{ESim-as-embedding} performs the best on Amazon. On the other four datasets, the average \textit{absolute} improvement of \textsc{Full} in comparison with \textsc{ESim-as-embedding} (resp., \textsc{Mp2v-as-embedding}) is 4.0\% (resp., 2.5\%) in terms of Micro F1. This finding demonstrates the advantage of our generation-guided embedding over meta-path based HIN embedding in this task. As mentioned above, ESim and Metapath2vec rely on meta-path guided random walks to model higher-order relationships between nodes. However, in our task, there is an evident generative process, with which we believe that characterizing generative relationships between various elements is more important than describing long-distance connections. 
(2) \textsc{Full} consistently outperforms the ablations ignoring different metadata, indicating that users, tags, products and contexts all play a positive role in classification. Meanwhile, their importance varies in different datasets. For example, product information is extremely important on Amazon according to Figure \ref{fig:embabl}(d). This is intuitive since Amazon's topic labels are essentially product categories. In contrast, user information is more useful on GitHub than it is on Amazon. This can be explained by the following statistics: GitHub-Bio has 351 pairs of documents sharing the same user, out of which 288 (82\%) have the same label; GitHub-AI has 348 pairs of repositories having the same user, among which 217 (62\%) belong to the same class.

\vspace{1mm}

\begin{figure}[t]
	\centering
	\subfigure[Micro F1]{
		\includegraphics[height=2.7cm]{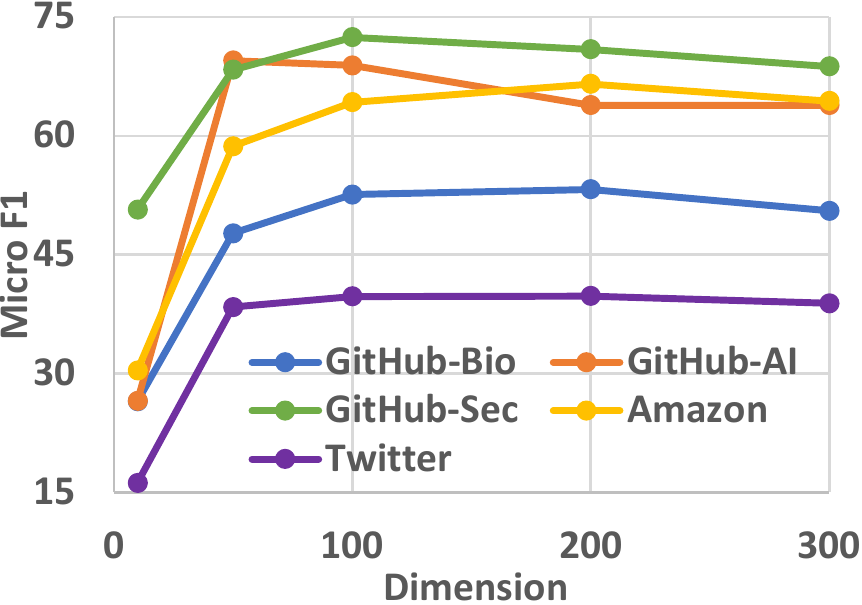}}
	\subfigure[Macro F1]{
		\includegraphics[height=2.7cm]{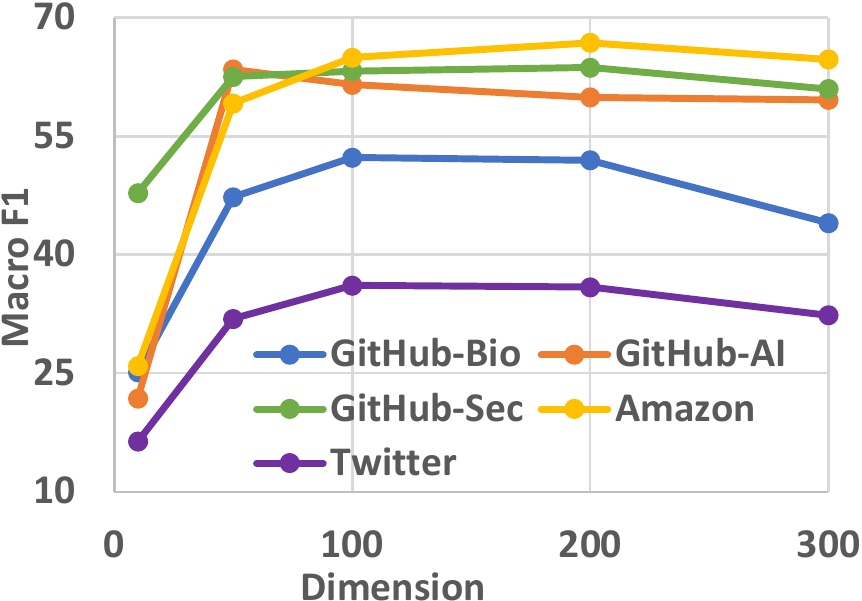}}
	\vspace{-1em}
	\caption{Performance of \textsc{MetaCat} with respect to the embedding dimension.} 
	\label{fig:dim}
\end{figure}

\noindent \textbf{Embedding Dimension.} Next, we investigate the performance of \textsc{MetaCat} with respect to its embedding dimension. Figure \ref{fig:dim} reports Micro and Macro F1 scores with 10, 50, 100, 200 and 300-dimensional embedding vectors. We can see that the performance drops when the dimension becomes too large. This observation is aligned with the results in \cite{tang2015line,shang2016meta}. In fact, too small dimension cannot sufficiently capture the semantics, while too large dimension may lead to some overfitting problems, especially under weak supervision. Figure \ref{fig:dim} shows that setting the dimension as 100 is reasonable in our experiments.

\vspace{1mm}

\begin{table}[]
\centering
\caption{Most similar words to a given label in the embedding space of \textsc{MetaCat}.}
\scalebox{0.81}{
\begin{tabular}{c|c|c}
	\hline
	Dataset                     & Label $l$              & Top similar words $w$                                                                                    \\ \hline
	\multirow{3}{*}{\textsc{GitHub-Bio}} & Genetics            & \begin{tabular}[c]{@{}c@{}}mega, ezmap, igess, \\ phenotypesimulator, multigems\end{tabular}          \\ \cline{2-3} 
	& \begin{tabular}[c]{@{}c@{}} Structural \\ Bioinformatics \end{tabular} & \begin{tabular}[c]{@{}c@{}}pywater, msisensor, knotty, \\ breakpointsurveyor, cmv\end{tabular}        \\ \hline
	\multirow{2}{*}{\textsc{GitHub-AI}}  & Entity Recognition  & ner, ld, conll, entity, tags                                                                          \\ \cline{2-3} 
	& Speech Synthesis    & tacotron, lecture, tts, lj, spectrogram                                                               \\ \hline
	\multirow{2}{*}{\textsc{Amazon}}     & Home \& Kitchen     & spinner, lettuce, salad, spinning, greens                                                             \\ \cline{2-3} 
	& Movies \& TV        & mafia, undercover, warren, depp, pacino                                                               \\ \hline
	\multirow{3}{*}{\textsc{Twitter}}    & Travel \& Transport & baggage, airway, claim, fdny, jetblue                                                                 \\ \cline{2-3} 
	& Shop \& Service     & \begin{tabular}[c]{@{}c@{}}keyfood, greenmarket, \\unsqgreenmarket, nailsbymii, nailart\end{tabular} \\ \hline
\end{tabular}
}
\label{table:case}
\vspace{-1em}
\end{table}

\noindent \textbf{Cases.} Table \ref{table:case} shows the top-5 similar words to a given label in the embedding space of \textsc{MetaCat}. Here the similarity between word $w$ and label $l$ is defined as $\text{cos}(\bme_w, \bme_l)$. Due to space limit, we demonstrate 8 categories from 4 datasets. We can find many strong topic indicators in Table \ref{table:case}. For example, in the GitHub domain, there are system/tool names (e.g., ``\textit{pywater}'' and ``\textit{tacotron}'') and dataset names (e.g., ``\textit{conll}'' and ``\textit{lj}'') related to the given category; in the Twitter domain, the top similar words mainly reflect the function of POIs (e.g., ``\textit{airway}'', ``\textit{jetblue}'' and ``\textit{unsqgreenmarket}'').

\subsection{Effect of Training Data Generation}
\label{sec:pseudoexp}
\begin{figure}[t]
	\centering
	\subfigure[Micro F1]{
		\includegraphics[height=2.7cm]{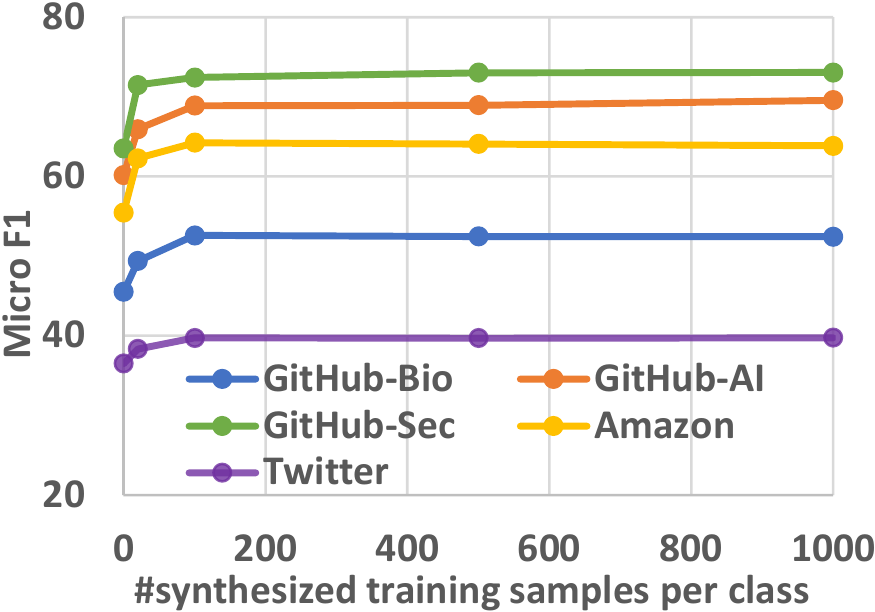}}
	\subfigure[Macro F1]{
		\includegraphics[height=2.7cm]{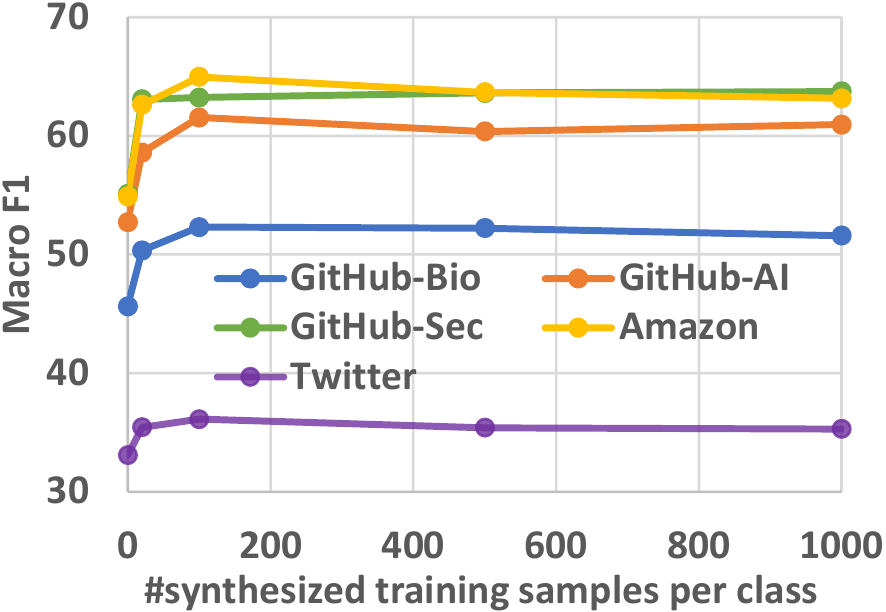}}
	\vspace{-1em}
	\caption{Performance of \textsc{MetaCat} with respect to the number of synthesized training samples.} 
	\vspace{-1em}
	\label{fig:pseudo}
\end{figure}

We generate 100 training samples for each class in all previous experiments. To investigate the effect of synthesized training data amount, we plot the performance of \textsc{MetaCat} with 0, 20, 100, 500 and 1000 generated samples per class in Figure \ref{fig:pseudo}.

When $|\mathcal{D}_l^*| < 100$, the F1 scores increase evidently with $|\mathcal{D}_l^*|$. For example, the average \textit{absolute} improvement of Micro F1 is 7.3\% on the five datasets when comparing $|\mathcal{D}_l^*| = 100$ with $|\mathcal{D}_l^*| = 0$ (i.e., using ``real'' training data only). This observation validates our claim that incorporating generated training data can boost classification performance. However, when the number of documents becomes larger, the performance change is quite subtle. In fact, the fluctuation is less than 1\% in most cases after $|\mathcal{D}_l^*| \geq 100$. Moreover, generating too many data will make the training process inefficient. In all, we believe having 100 to 500 synthesized training samples per class will strike a good balance in our task.

\subsection{Effect of ``Real'' Training Data}
\begin{figure}[t]
	\centering
	\subfigure[\textsc{GitHub-Sec}]{
		\includegraphics[height=2.7cm]{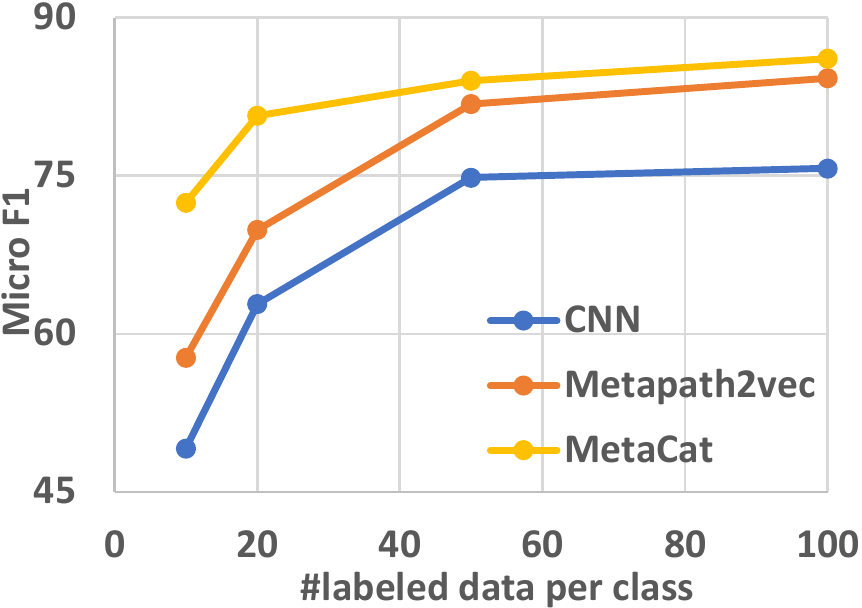}}
	\subfigure[\textsc{Amazon}]{
		\includegraphics[height=2.7cm]{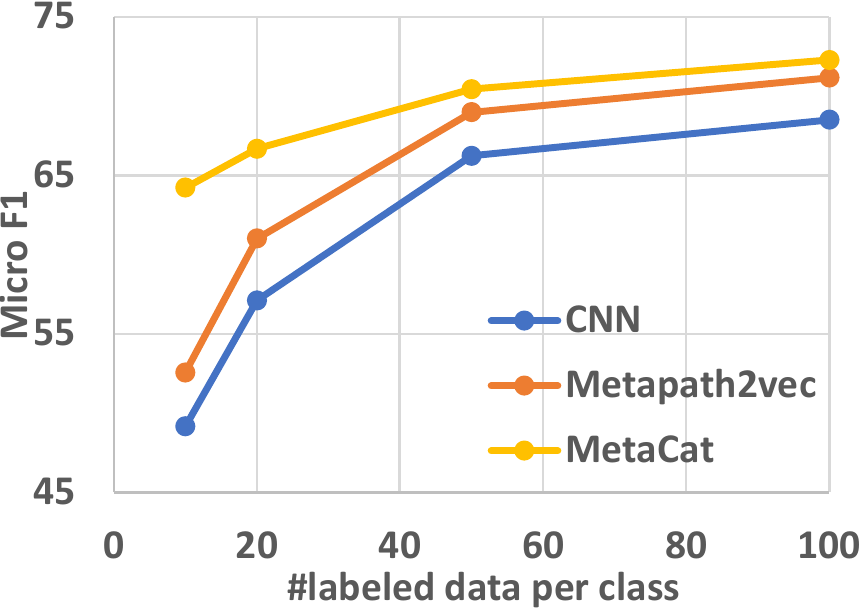}}
	\subfigure[Comparisons of the three approaches]{
		\includegraphics[width=0.46\textwidth]{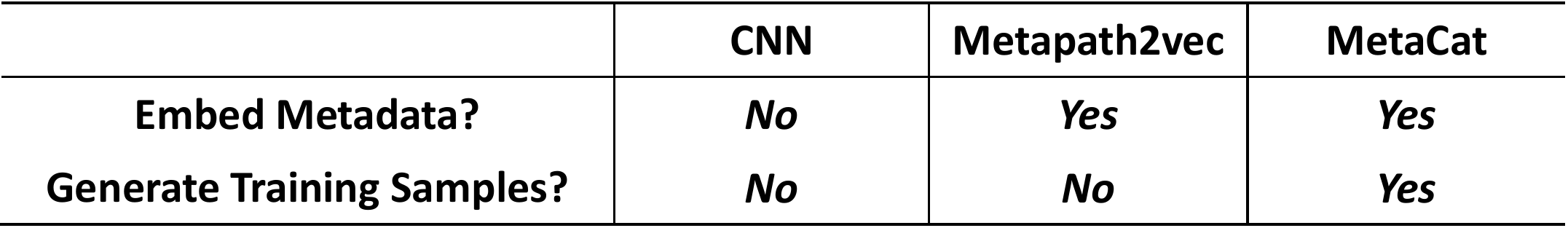}}
	\vspace{-1em}
	\caption{Micro F1 scores of CNN, Metapath2vec and \textsc{MetaCat} with respect to the number of ``real'' training samples.} 
	\vspace{-1em}
	\label{fig:label}
\end{figure}
According to Sections \ref{sec:hin} and \ref{sec:pseudoexp}, we already know both embedding and generation modules are helpful when supervision is minimal. Now we examine their effects as ``real'' training data is increased. Three approaches are picked: CNN (does not embed metadata; does not generate training samples), Metapath2vec (embeds metadata; does not generate training samples) and \textsc{MetaCat} (embeds metadata; generates training samples). Figure \ref{fig:label} shows their performance with 10, 20, 50 and 100 ``real'' training samples per class on \textsc{GitHub-Sec} and \textsc{Amazon}.

Let us first compare \textsc{MetaCat} and Metapath2vec. When the amount of labeled data is relatively large, \textsc{MetaCat} only outperforms Metapath2vec by a small margin. However, if fewer labeled documents are provided, Metapath2vec exhibits more evident performance drop. For example, when $|\mathcal{D}_l|=10$, the gap between \textsc{MetaCat} and Metapath2vec is 14.7\% on GitHub-Sec; when $|\mathcal{D}_l|=100$, the gap becomes 1.8\%. This observation shows that training data generation is more powerful when supervision is weaker.

Then we proceed to the curves of CNN. Similar to Metapath2vec, its performance drops drastically when $|\mathcal{D}_l|$ becomes small. However, there is always a large gap between CNN and \textsc{MetaCat}. When labeled training samples are sufficient, this gap is mainly attributed to the usage of metadata in embedding (because we already show that training data generation plays a limited role in this case); when there is less supervision, the gap becomes a composite effect of utilizing metadata and generating synthesized samples.
Different from training data generation, no matter how many labeled documents are involved, exploiting metadata in embedding is always helpful.

\section{Related Work}
\noindent \textbf{Weakly Supervised Text Classification.}
Two forms of supervisions are commonly studied under weakly supervised text classification: (1) \textit{class-related keywords}. For example, dataless classification \cite{chang2008importance} only relies on descriptive keywords and has no requirement of any labeled document. Subsequent studies either extend topic models \cite{lu2008opinion,chen2015dataless,li2016effective,li2018dataless} or exploit embedding techniques \cite{li2018unsupervised,meng2020cate} and contextualized representation learning models \cite{mekala2020contextualized} to incorporate such keyword information. (2) \textit{a small set of labeled documents}. For example, \cite{tang2015pte} jointly learns word/document/label representations; \cite{meng2018weakly} generates pseudo documents to help the training process; \cite{yao2019graph} applies graph convolutional networks on the word-document co-occurrence graph.
\cite{meng2019weakly} and \cite{xiao2019efficient} further study weakly supervised hierarchical classification. However, these studies focus on text without metadata, which restricts their capacity in some practical scenarios. In contrast, \textsc{MetaCat} goes beyond plain text classification and utilizes multi-modal signals.

\vspace{1mm}

\noindent \textbf{Text Classification with Metadata.} Existing studies apply metadata to improve the performance of a text classifier, such as user and product information in sentiment analysis \cite{tang2015learning}, author information in paper topic classification \cite{rosen2004author}, and user biography data in tweet localization \cite{zhang2017rate}. However, each of these frameworks focuses on one specific type of data, while we propose a general framework to deal with various data sources and metadata types. Kim et al. \cite{kim2019categorical} study how to ``inject'' categorical metadata information into neural text classifiers. Their model can be applied in many cases such as review sentiment classification and paper acceptance classification. However, the model is designed under fully supervised settings and does not tackle label scarcity.

\vspace{1mm}

\noindent \textbf{Embedding Learning of Heterogeneous Data.}
Heterogeneous Information Network (HIN) embedding is the most common technique to encode multi-modal signals. Many HIN embedding methods \cite{dong2017metapath2vec,shang2016meta,fu2017hin2vec} leverage meta-path guided random walks to jointly model multiple interactions in a latent embedding space. One can refer to a recent tutorial \cite{shi2019recent} for more related studies. We have discussed the differences between HIN embedding and our generation-guided embedding in Section \ref{sec:embedding}. From the view of applications, several studies apply HIN embeddings into downstream classification tasks such as malware detection \cite{hou2017hindroid} and medical diagnosis \cite{hosseini2018heteromed}. Both \cite{hou2017hindroid} and \cite{hosseini2018heteromed}, as well as experiments in \cite{dong2017metapath2vec,shang2016meta,fu2017hin2vec}, deal with structured data under fully supervised settings.

\section{Conclusions and Future Work}
We presented \textsc{MetaCat}, a framework to categorize text with metadata under minimal supervision. To tackle the challenges of data heterogeneity and label scarcity, we propose a generation-guided embedding module and a training data generation module. Both modules are derived from a generative process characterizing the relationships between text and metadata. We demonstrate the effectiveness of \textsc{MetaCat} on five datasets. Moreover, we validate the design of our framework by showing (1) the superiority of our embedding module towards meta-path based HIN embedding methods and (2) the significant contribution of our generation module especially when the amount of supervision is very limited. 

For future work, first, it is interesting to study how to effectively integrate different forms of supervision (e.g., annotated documents and class-related keywords) to further boost the performance. Second, we would like to explore the possibility of coupling heterogeneous signal embedding and graph neural networks during the classification process. 

\begin{acks}
We thank Sha Li for useful discussions. The research was sponsored in part by DARPA under Agreements No. W911NF-17-C-0099 and FA8750-19-2-1004, National Science Foundation IIS 16-18481, IIS 17-04532, and IIS-17-41317, and DTRA HDTRA11810026. 
Any opinions, findings, and conclusions or recommendations expressed in this document are those of the author(s) and should not be interpreted as the views of any U.S. Government. The U.S. Government is authorized to reproduce and distribute reprints for Government purposes notwithstanding any copyright notation hereon.
We thank anonymous reviewers for valuable and insightful feedback.
\end{acks}

\balance
\bibliography{sigir}
\end{spacing}

\end{document}